\documentclass[a4paper,10pt,margin=0.5in]{extarticle}
\usepackage{geometry}
\usepackage[english]{babel}
\usepackage[utf8]{inputenc}
\usepackage{amsmath}
\usepackage{float}
\usepackage{graphicx,subfigure}
\usepackage[utf8]{inputenc}
\usepackage[english]{babel}
\usepackage{amsfonts}
\usepackage{comment} 
\usepackage{tikz}
\usepackage{mathtools}
\usepackage{fixltx2e}
\usepackage{pifont}
\usepackage{caption}
\usepackage[colorinlistoftodos]{todonotes}

\usepackage{amsmath,amsfonts,amssymb,amsthm,epsfig,epstopdf,array}

\newtheorem{assumption}{Assumption}

\usepackage{caption}

\newcommand{\norm}[1]{\left\lVert#1\right\rVert}
\usepackage{verbatim}
\usepackage{balance}

\usepackage{enumitem}
\usepackage[bottom]{footmisc}
\usepackage[utf8]{inputenc}
\usepackage[english]{babel}

\usepackage[linkcolor=blue,colorlinks=true]{hyperref}
\usepackage{color, colortbl}
\author{Jaskaran Grover$^{1}$, Changliu Liu$^{1}$, Katia Sycara $^{1}$
	\thanks{$^{1}$J. Grover, C. Liu, K. Sycara are with the Robotics Institute at
		Carnegie Mellon University, 5000 Forbes Avenue, Pittsburgh, PA 15213, USA.
		{\tt\small \{jaskarag,cliu6,sycara\}@andrew.cmu.edu}}}%
\begin{document}
	\title{\LARGE \bf
		Feasible Region-based Identification Using Duality (Extended Version) }
	\date{}
	\maketitle
	\begin{abstract}
We consider the problem of estimating bounds on parameters representing tasks being performed by individual robots in a multirobot system. In our previous work, we derived necessary conditions based on persistency of excitation analysis for exact identification of these parameters. We concluded that depending on the robot's task, the dynamics of individual robots may fail to satisfy these conditions, thereby preventing exact inference. As an extension to that work, this paper focuses on estimating bounds on task parameters when such conditions are not satisfied. Each robot in the team uses optimization-based controllers for mediating between task satisfaction and collision avoidance. We use KKT conditions of this optimization and SVD of active collision avoidance constraints to derive explicit relations between Lagrange multipliers, robot dynamics and task parameters. Using these relations, we are able to derive bounds on each robot's task parameters. Through numerical simulations we show how our proposed region based identification approach generates feasible regions for parameters when a conventional estimator such as a UKF fails. Additionally, empirical evidence shows that this approach generates contracting sets which converge to the true parameters much faster than the rate at which a UKF based estimate converges. Videos of these results are available at \url{https://bit.ly/2JDMgeJ}
\end{abstract}
\section{Introduction}
Task-based motion planning and control for multiple robots has several applications including search and rescue \cite{kantor2003distributed}, sensor coverage \cite{cortes2004coverage} and environmental exploration \cite{burgard2005coordinated}. The resiliency and robustness that can be achieved by multiple robots is superior to that achievable by one \cite{ogren2002control}, \cite{olfati2007consensus}. While controlling these robots can be treated as a forward problem, in this paper we consider the inverse problem  of task inference  \cite{byravan2015graph}, which has several applications of its own. For example, analyzing how easy it is to infer tasks being performed by a team of robots by simply observing them can reveal vulnerabilities of that robot team to adversarial attacks. This analysis can be used to design privacy preserving controllers to make the team resilient to attacks. In this paper, we take the view of an observer monitoring a team of robots in which each robot is performing some task. The question we want to address is how can the observer compute \textit{reasonable bounds} on the parameters representing the task of each robot by measuring their positions over time. 

The approach to infer parameters by estimating bounds is different from inferring them using point-wise estimation algorithms such as an Unscented Kalman Filter (UKF) \cite{thrun2002probabilistic} or adaptive observers \cite{na2015robust},\cite{adetola2010performance},\cite{hartman2012robust}. These algorithms are point-wise estimators because they generate a single estimate of the parameter that converges to the true parameter. Their convergence, however, is only guaranteed when the system dynamics satisfy the \textit{persistency of excitation} condition \cite{willems2005note}. When an observer has the freedom to command inputs to the system, he can design \textit{sufficiently rich} inputs that will ensure convergence of these estimators to the true parameters of the system. This open-loop identification has been used for identifying physical parameters of manipulators \cite{yang2018adaptive} and quadrotors \cite{zhao2018online}. 

By contrast, our scenario is different from these systems because of two reasons. Firstly, our observer can only observe the robots but not intervene physically. Hence the parameter identification must occur in a closed-loop way.  Secondly, while manipulators and quadrotors are monolithic, our  ``plant" is comprised of multiple robots. In our previous work \cite{grover2020parameter}, we derived necessary identifiability conditions for this system and showed that multiple active interactions between robots leads to violation of the persistency of excitation condition. Since interactions between robots are inevitable, point-wise parameter estimation algorithms would fail to converge owing to the violation of this condition.

This takeaway from our previous work forms the motivation for this paper. That is, while it may not be possible to get an exact estimate of the task parameter of a robot, we want to determine if we can estimate any bounds for it. To address this problem, we propose a \textit{feasible-region based parameter identification} algorithm that generates a bounded set where the true task parameter of each robot must belong. Further, we design this algorithm in a way such that the ``measure" of this set (\textit{i.e.} its size) is non-increasing with time. There are several advantages of our proposed identification approach over point-wise estimation. Firstly, this algorithm is anytime \textit{i.e.} the set computed at a given time is comprised of parameters that are all valid candidates for explaining the robot measurements observed until that time. Secondly, this algorithm does not have any gains to tune unlike Kalman filters or adaptive observers \cite{hartman2012robust} where tuning relies on user's experience.  Thirdly, this algorithm can also work symbiotically with point-wise estimation algorithms and can expedite their convergence by continually projecting their estimates to the feasible regions generated by this algorithm. Finally, our empirical evidence suggests that even in cases where point-wise estimation is possible, region-based estimation converges much faster than a point-wise estimator (we have used a UKF for comparison).

The outline of this paper is as follows. In section \ref{Approaches_for_Parameter_Identification}, we give a brief review of point-wise parameter estimation following the development in \cite{grover2020parameter,adetola2008finite} and establish notation for our proposed \textit{feasible region-based} identification approach. In section \ref{ControlReview}, we discuss how prior work \cite{wang2017safety} has addressed the multirobot task satisfaction problem using optimization-based controllers. Although prior work has mostly considered the task of making robots reach some goals, we make the formulation abstract to allow for arbitrary tasks.  We formalize the \textit{feasible region-based} task identification problem in section \ref{ProblemFormulation}.  The main contributions begin from section \ref{KKT Conditions} where we derive the KKT conditions of the control-synthesis optimization.  These conditions allow us to pose an equality constrained optimization problem (EQP) formulated using the set of active interactions (\textit{i.e.} active constraints) of the ego robot. In section \ref{SVD}, we classify this robot's dynamics based on the number of constraints in this EQP, and linear independence relations amongst these constraints. Taking the SVD of these constraints allows us to derive explicit relations between robot dynamics, Lagrange multipliers and task parameters. Finally, by using these relations along with the KKT conditions, we derive analytical descriptions of the sets where the task parameters must belong as we wanted. We demonstrate the power of this identification approach through numerical simulations in section \ref{Results}. In these simulations, we consider geometric settings that span all combinations of number of active constraints and linear independence relations we theorized in section \ref{SVD}. For these cases, we show through simulations how the bounds on task parameters computed by our approach converge much faster to the true parameter than an Unscented Kalman Filter. We conclude in \ref{Conclusions} by summarizing and provide directions for future work.

\section{Approaches for Parameter Identification}
\label{Approaches_for_Parameter_Identification}
We describe two different approaches for parameter identification. These include (a) Point-wise identification and (b) Feasible-region based identification. The need for distinguishing between these two approaches is necessitated by our application \textit{i.e.} task inference for multirobot systems. For these systems, we will demonstrate that approach (a) which is the de-facto standard for parameter identification, can often fail to converge owing to unavoidable interactions between robots. In such circumstances, approach (b) on the other hand, can provide reasonable bounds on task parameters, if not their exact values.
\subsection{Point-wise identification}
The \textit{point-wise identification} approach focuses on designing an online parameter update law which ensures that the estimate of the parameter converges to the true parameter of the underlying system. Several known estimation algorithms fall in this category, including RLS \cite{edgar2010recursive}, UKF, EKF \cite{thrun2002probabilistic} and the ones based on adaptive observers \cite{adetola2008finite,adetola2010performance,na2015robust}. Adaptive observer based estimators provide conditions based on the \textit{persistency of excitation} criterion \cite{ioannou2012robust,narendra2012stable} under which convergence to the true parameters is achieved. To formalize point-wise identification, consider a nonlinear system as follows
\begin{align}
\label{original}
    \dot{\boldsymbol{x}}=G(\boldsymbol{x})\boldsymbol{\theta} + f(\boldsymbol{x}),
\end{align}
where $\boldsymbol{x} \in \mathbb{R}^n$ is the measurable state, $\boldsymbol{\theta}\in\mathbb{R}^p$ is the unknown parameter and $G(\boldsymbol{x}):\mathbb{R}^{n} \longrightarrow \mathbb{R}^{n \times p}$, $f(\boldsymbol{x}):\mathbb{R}^{n} \longrightarrow\mathbb{R}^{n}$ are known functions. For example, in our context, $\boldsymbol{x}(t)$ will correspond to the position of the ego robot under observation and $\boldsymbol{\theta}$ denotes the task parameters of its controller that the observer wishes to infer. We can safely assume that the observer runs several parallel estimators, one for estimating the parameters of each robot, so the observer's focus is on the ego robot. 
Adhering to the \textit{pointwise identification} paradigm, the observer's problem is to design an estimation law $\dot{\hat{\boldsymbol{\theta}}}=\psi(\hat{\boldsymbol{\theta}},\boldsymbol{x})$ that guarantees convergence of $\hat{\boldsymbol{\theta}} \longrightarrow \boldsymbol{\theta}$ by using $\boldsymbol{x}(t)$ over some $t \in[0,T]$ where $T$ is large enough.  
\subsection{Feasible region-based identification}
\label{TypeBidentification}
This approach computes a set in the parameter space where the parameter of the underlying system must belong. The observer may use some measurements $\boldsymbol{y}(t)$ and compute a set $\boldsymbol{\Theta}(t) \subset \mathbb{R}^p$
such that 
\begin{enumerate}
    \item $\boldsymbol{\Theta}(t)$ is bounded.
    \item $\boldsymbol{\theta} \in \boldsymbol{\Theta}(t)$ $\forall t$.
    \item The ``measure" of this set, denoted by $\texttt{Vol}(\boldsymbol{\Theta}(t))$ is non-increasing with time. 
\end{enumerate}
Ideally, if condition (2) is satisfied and condition (3) is replaced with a strict decrease in $\texttt{Vol}(\boldsymbol{\Theta}(t))$, it is guaranteed that $\texttt{lim}_{t \rightarrow \infty}\boldsymbol{\Theta}(t) =\boldsymbol{\theta}$. However, the relaxed condition (3) \textit{i.e.} $\texttt{Vol}(\boldsymbol{\Theta}(t))$ is non-increasing is easier to ensure in practice and will be our focus. 
One way to define a set that satisfies these three conditions is as follows. Suppose at a given time $t$, we deduce that $\boldsymbol{\theta} \in \boldsymbol{\Omega}(t)$ where
\begin{align}
\label{setdefinition}
    \boldsymbol{\Omega}(t) \coloneqq \{\boldsymbol{\hat{\theta}}\in \mathbb{R}^p \vert \boldsymbol{g}(\boldsymbol{y}(t),\boldsymbol{\hat{\theta}}) \prec \boldsymbol{0}\},
\end{align}
for some function $\boldsymbol{g}(\boldsymbol{y}(t),\boldsymbol{\hat{\theta}})$ that depends on the observer's measurements $\boldsymbol{y}(t)$. Note that $\boldsymbol{\Omega}(t)$ need not be bounded, that would depend on how $\boldsymbol{g}(\boldsymbol{y}(t),\boldsymbol{\hat{\theta}})$ is defined. Further, suppose $\boldsymbol{\Theta}_0 \subset \mathbb{R}^p$ is a known time-invariant compact set in which $\boldsymbol{\theta}$ is known to belong a-priori. Then defining 
\begin{align}
\label{timeintersection}
    \boldsymbol{\Theta}(t) \coloneqq \bigcap_{0\leq \tau\leq t}\boldsymbol{\Omega}(\tau) \cap \boldsymbol{\Theta}_0
\end{align}
will ensure satisfaction of the three  conditions proposed above. Indeed, let $t_1$ and $t_2$ be two time instants such that $t_1 < t_2$, then from \eqref{timeintersection} it is evident that $\boldsymbol{\Theta}(t_2)\subseteq \boldsymbol{\Theta}(t_1)$. This would in-turn imply that
$\texttt{Vol}(\boldsymbol{\Theta}(t_2)) \leq \texttt{Vol}(\boldsymbol{\Theta}(t_1))$ or in other words, $\texttt{Vol}(\boldsymbol{\Theta}(t))$ non-increasing.  Since  $\boldsymbol{\Theta}(t)$ is derived from the set $\boldsymbol{\Omega}(t)$, it suffices to focus on constructing the \textit{instantaneous feasible set} $\boldsymbol{\Omega}(t)$ \textit{i.e.} deriving $\boldsymbol{g}(\boldsymbol{y}(t),\boldsymbol{\hat{\theta}})$.

In the rest of the paper, our goal will be to design a \textit{feasible-region based identification} method for inferring task parameters of a multirobot system. To construct the the \textit{instantaneous feasible set} $\boldsymbol{\Omega}(t)$, we need to define functions $\boldsymbol{g}(\boldsymbol{y}(t),\boldsymbol{\hat{\theta}})$ needed in \eqref{setdefinition}. This requires some assumptions on the controllers that the robots use to perform their tasks (parametrized by $\boldsymbol{\theta}$) and the resulting dynamics of the robots. The most basic of these assumptions is safety \textit{i.e.} we assume that all robots have an underlying collision avoidance mechanism to ensure collision-free motions while they perform their tasks. To mediate between task satisfaction and safety, a common approach is to use reactive optimization-based controllers as has been demonstrated in \cite{wang2017safety,grover2019deadlock}. Throughout the rest of the paper, we will assume that the robots use such controllers. The identification framework we develop assumes \textit{Control Barrier Function based QPs} (CBF-QPs) as the underlying controllers, just to illustrate the mechanics of constructing $\boldsymbol{\Omega}(t)$. However, our approach can be used to construct this set for any other type of optimization-based controller as well \cite{wei2019safe}. 
\section{Task Satisfaction and Collision Avoidance}
\label{ControlReview}
We give a brief summary of CBF-QPs here, referring the reader to \cite{wang2017safety} for a detailed treatment. Let there be a total of $M+1$ robots in the system. From the perspective of an ego robot, the remaining $M$ robots are ``cooperative obstacles" who share the responsibility of avoiding collisions with the ego robot, while performing their own respective tasks. In the following discussion, the focus is on the ego robot. This robot follows single-integrator dynamics \textit{i.e.}
\begin{align}
\dot{\boldsymbol{x}} = \boldsymbol{u},
\end{align}
where $\boldsymbol{x}=(p_x,p_y)\in\mathbb{R}^2$ is its position and $\boldsymbol{u}\in\mathbb{R}^2$ is its velocity (\textit{i.e.} the control input). Assume that this robot can use a nominal controller $\hat{\boldsymbol{u}}(\boldsymbol{x})= C(\boldsymbol{x})\boldsymbol{\theta} + d(\boldsymbol{x})$ that guarantees satisfaction of the task when there is no other robot in the system. Here, $\boldsymbol{\theta}$ is the task-parameter the observer wishes to infer, and $C(\boldsymbol{x}),d(\boldsymbol{x})$ are appropriate functions. As an example, if the ego robot's task is to reach a goal at $\boldsymbol{x}_d$, the ego robot can use  $\hat{\boldsymbol{u}}(\boldsymbol{x})=-k_p(\boldsymbol{x}-\boldsymbol{x}_d)$, a proportional controller that guarantees $\boldsymbol{x} \rightarrow \boldsymbol{x}_d$. If the observer's problem is to infer the goal $\boldsymbol{x}_d$, then defining $C(\boldsymbol{x})=k_p$, $d(\boldsymbol{x})=-k_p\boldsymbol{x}$ and $\boldsymbol{\theta}=\boldsymbol{x}_d$ gives $C(\boldsymbol{x})\boldsymbol{\theta} + d(\boldsymbol{x})=-k_p(\boldsymbol{x}-\boldsymbol{x}_d)$.

Now although $\hat{\boldsymbol{u}}(\boldsymbol{x})$ by itself ensures task completion, the robots must also avoid collisions amongst themselves. To that end, the ego robot must maintain a distance of at-least $D_s$ with all the other robots located at $\{\boldsymbol{x}^o_j\}_{j=1}^{M}$. That is the their positions $(\boldsymbol{x},\boldsymbol{x}^o_j)$ must satisfy $ \norm{\Delta \boldsymbol{x}_{j}}^2\geq D_s^2$ where $\Delta \boldsymbol{x}_{j} \coloneqq \boldsymbol{x}-\boldsymbol{x}^o_j$.  To combine the  collision avoidance requirement with the task-satisfaction objective, the ego robot solves a QP that computes a controller closest to the nominal task control $\hat{\boldsymbol{u}}(\boldsymbol{x})$ and satisfies $M$ collision avoidance constraints as follows:
\begin{align}
\label{optimization_formulation_1_static_obstacles}
\begin{aligned}
\boldsymbol{u}^*&= \underset{\boldsymbol{u}}{\arg\min}
& & \norm{\boldsymbol{u} - \hat{\boldsymbol{u}}(\boldsymbol{x})}^2 \\
& \text{subject to}
& & A(\boldsymbol{x})\boldsymbol{u} \leq \boldsymbol{b}(\boldsymbol{x})  
\end{aligned}
\end{align}
Here $A(\boldsymbol{x})\in \mathbb{R}^{M\times 2}$, $\boldsymbol{b}(\boldsymbol{x})\in \mathbb{R}^{M}$ are defined such that the $j^{th}$ row of $A$ is $\boldsymbol{a}^T_j$ and the $j^{th}$ element of $\boldsymbol{b}$ is $b_j$:
\begin{align}
\label{Ab_static}
\boldsymbol{a}^T_j(\boldsymbol{x}) &\coloneqq-\Delta \boldsymbol{x}^T_{j}=-(\boldsymbol{x}-\boldsymbol{x}^o_j)^T \nonumber \\
b_j(\boldsymbol{x}) &\coloneqq \frac{\gamma}{2} (\norm{\Delta \boldsymbol{x}_{j}}^2-D_s^2) \mbox{   }\forall j \in \{1,2,\dots,M\}
\end{align}
The ego robot solves this QP at every time step to determine its control $\boldsymbol{u}^*$, which ensures collision avoidance while encouraging satisfaction of the task parametrized by $\boldsymbol{\theta}$.

Aside from its position $\boldsymbol{x}$, the ego robot's control $\boldsymbol{u}^*$ depends on task parameter $\boldsymbol{\theta}$. This dependence is implicitly encoded through the cost function of  \eqref{optimization_formulation_1_static_obstacles} (recall $\hat{\boldsymbol{u}}(\boldsymbol{x})=C(\boldsymbol{x})\boldsymbol{\theta} + d(\boldsymbol{x})$). To highlight this dependence, we denote the control as $\boldsymbol{u}^*_{\boldsymbol{\theta}}(\boldsymbol{x})$ where $\boldsymbol{\theta}$ is the unknown parameter the observer aims to identify.  Next, we formulate the problem that the observer seeks to solve.
\section{Region-based Identification Problem Formulation}
\label{ProblemFormulation}
The observer's problem is to identify a region in the parameter space consistent with conditions (1) and (2) of the \textit{feasible-region based identification} approach proposed in Sec. \ref{Approaches_for_Parameter_Identification}. That is, the observer must identify a set $\boldsymbol{\Theta}(t) \subset \mathbb{R}^p$ where $\boldsymbol{\theta}$ must lie $\forall$ $t\in [0,T]$ such that its volume is non-increasing over this duration. 
Following the analysis in subsec. \ref{TypeBidentification}, it suffices to estimate the \textit{instantaneous feasible set} $\boldsymbol{\Omega}(t)$ \eqref{setdefinition} as the 0-sublevel set of a function $\boldsymbol{g}(\boldsymbol{y}(t),\boldsymbol{\hat{\theta}})$.
The measurements $\boldsymbol{y}(t)$ \eqref{setdefinition} that the observer can use to compute this set include (a) the positions of the ego robot $\boldsymbol{x}(t)$ and (b) the positions of the ``cooperative obstacles"   $\boldsymbol{x}^o_j(t) \mbox { } \forall j \in \{1,2,\cdots,M\}$ over $t \in [0,T]$ \textit{i.e.} $\boldsymbol{y}(t)=(\boldsymbol{x}(t),\boldsymbol{x}^o_1(t),\cdots,\boldsymbol{x}^o_M(t))$. The observer will run $M+1$ parallel set-estimators to compute $\{\boldsymbol{\Theta}^i(t)\}_{i=1}^{M+1}$ for each robot in the team since each robot has its unique task parameter $\boldsymbol{\theta}_i$.

Focusing on the ego robot, the observer needs to derive functions $\boldsymbol{g}(\boldsymbol{y}(t),\hat{\boldsymbol{\theta}})$ needed in \eqref{setdefinition} to compute $\boldsymbol{\Omega}(t)$. For that,
the observer must know an explicit relation between the dynamics of the ego robot and the parameters $\boldsymbol{\theta}$ of the form $\dot{\boldsymbol{x}}=G(\boldsymbol{x})\boldsymbol{\theta} + f(\boldsymbol{x})$ similar to \eqref{original}. That is, it must know  $G(\boldsymbol{x})$ and $f(\boldsymbol{x})$. However, owing to the fact that $\dot{\boldsymbol{x}}=\boldsymbol{u}^*_{\boldsymbol{\theta}}(\boldsymbol{x})$ is optimization-based \eqref{optimization_formulation_1_static_obstacles}, such explicit relations are not known. It is possible to derive these relations by focusing on the KKT conditions of \eqref{optimization_formulation_1_static_obstacles}. We derive these conditions in the next section using which the observer will be able to compute the set $\boldsymbol{\Omega}(t)$. Before proceeding, let's state all the assumptions on the observer's knowledge:
\begin{assumption}
\label{ass3}
The observer can measure both the position $\boldsymbol{x}(t)$ and velocity (\textit{i.e.} the control $\boldsymbol{u}^*_{\boldsymbol{\theta}}(\boldsymbol{x}(t))$ of the ego robot.
\end{assumption}
\begin{assumption}
	\label{ass1}
	The observer knows the form of safety constraints $A(\boldsymbol{x}),\boldsymbol{b}(\boldsymbol{x})$ in \eqref{optimization_formulation_1_static_obstacles} and that the cost function is of the form $\norm{\boldsymbol{u} - \hat{\boldsymbol{u}}(\boldsymbol{x})}^2$.
\end{assumption}
\begin{assumption}
\label{ass2}
The observer knows the task functions $C(\boldsymbol{x}),d(\boldsymbol{x})$ of $\hat{\boldsymbol{u}}(\boldsymbol{x})=C(\boldsymbol{x})\boldsymbol{\theta}+d(\boldsymbol{x})$
\end{assumption}
Assumption \ref{ass3} is not restrictive because positions are easily measurable and velocities can be obtained through numerical differentiation of positions.
Assumption \ref{ass1} is needed since we are interested in deriving expressions of $\boldsymbol{g}(\boldsymbol{y}(t),\hat{\boldsymbol{\theta}})$ and that requires knowledge of  the underlying control synthesis approach.  Assumptions \ref{ass2} is not restrictive in practice because the observer can hypothesize on the high-level task by observing the robots and query a library of task-to-function mappings for guessing $C(\boldsymbol{x}),d(\boldsymbol{x})$. 
\section{Analysis using KKT conditions}
\label{KKT Conditions}
To analyze the relation between the optimizer of \eqref{optimization_formulation_1_static_obstacles}  \textit{i.e.} $\boldsymbol{u}^*_{\boldsymbol{\theta}}(\boldsymbol{x})$ and parameters $\boldsymbol{\theta}$, we look at the KKT conditions of this QP. These  are necessary and sufficient conditions satisfied by $\boldsymbol{u}^*_{\boldsymbol{\theta}}(\boldsymbol{x})$. The Lagrangian for \eqref{optimization_formulation_1_static_obstacles} is
\begin{align}
L(\boldsymbol{u},\boldsymbol{\mu}) =  \norm{\boldsymbol{u} - \hat{\boldsymbol{u}}}^2_2  + \boldsymbol{\mu}^T(A\boldsymbol{u}-\boldsymbol{b}) \nonumber.
\end{align}
Let $(\boldsymbol{u}^*_{\boldsymbol{\theta}},\boldsymbol{\mu}^*_{\boldsymbol{\theta}})$ be the optimal primal-dual solution to  \eqref{optimization_formulation_1_static_obstacles}. The KKT conditions are \cite{boyd2004convex}:
\begin{enumerate}
	\item Stationarity: $\nabla_{\boldsymbol{u}}L(\boldsymbol{u},\boldsymbol{\mu})\vert_{(\boldsymbol{u}^*_{\boldsymbol{\theta}},\boldsymbol{\mu}^*_{\boldsymbol{\theta}})} = 0$,
	\begin{align}
	\label{stationarity1}
		\implies \boldsymbol{u}^*_{\boldsymbol{\theta}} &= \hat{\boldsymbol{u}} - \frac{1}{2}\sum_{j \in \{1,\cdots,M\}}\mu^*_{j{\boldsymbol{\theta}}}\boldsymbol{a}_{j}
	\nonumber \\
    &= \hat{\boldsymbol{u}} - \frac{1}{2} A^T \boldsymbol{\mu}^*_{\boldsymbol{\theta}} .
	\end{align}
\item Primal Feasibility 
\begin{align}
\label{primal_feasibility1}
A\boldsymbol{u}^*_{\boldsymbol{\theta}}\leq \boldsymbol{b} \iff \boldsymbol{a}^T_{j}\boldsymbol{u}^*_{\boldsymbol{\theta}} \leq b_{j} \mbox{ }\forall j \in \{1,\cdots,M\}.
\end{align}
\item Dual Feasibility 
\begin{align}
\label{dual_feasibility1}
{\mu^*_{j{\boldsymbol{\theta}}}} \geq 0 \mbox{  }	 \forall j \in \{1,2,\cdots,M\}.
\end{align}
\item Complementary Slackness 
\begin{align}
	\label{complimentarty slackness1}
	\mu^*_{j{\boldsymbol{\theta}}} \cdot (\boldsymbol{a}^T_{j}\boldsymbol{u}^*_{\boldsymbol{\theta}} -b_{j}) = 0 
	 \mbox{   }\forall j \in \{1,2,\cdots,M\}.
\end{align}
\end{enumerate}
We define the set of active and inactive constraints as
\begin{align}
\label{activeinactive}
	\mathcal{A}(\boldsymbol{u}^*_{\boldsymbol{\theta}}) \coloneqq \{j \in \{1,2,\cdots,M\} \mid \boldsymbol{a}^T_{j}\boldsymbol{u}^*_{\boldsymbol{\theta}} = b_{j} \}, \\
	\mathcal{IA}(\boldsymbol{u}^*_{\boldsymbol{\theta}}) \coloneqq \{j \in \{1,2,\cdots,M\} \mid \boldsymbol{a}^T_{j}\boldsymbol{u}^*_{\boldsymbol{\theta}} < b_{j} \} .
\end{align}
The set of active constraints qualitatively represents those robots that the ego robot ``worries" about for collisions. From the perspective of the ego robot, we simply refer to the ``other robots" as \textit{obstacles}.  
\subsection{Using KKT conditions for defining the feasible region:}
\label{KKTFeasible}
Now we briefly describe how these conditions are useful for the computing the \textit{instantaneous feasible set} where the parameter $\boldsymbol{\theta}$ can lie \textit{i.e.} the set $\boldsymbol{\Omega}(t)$ defined in \eqref{setdefinition}.  
\begin{enumerate}
    \item \textbf{Inactive constraints:} \label{inactiveconstraints} From the definition of inactive constraints \eqref{activeinactive}, recall $\boldsymbol{a}^T_{j}\boldsymbol{u}^*_{\boldsymbol{\theta}} < b_{j}$ $\forall j \in \mathcal{IA}(\boldsymbol{u}^*_{\boldsymbol{\theta}})$. Thus, deriving an explicit expression for $\boldsymbol{u}^*_{\boldsymbol{\theta}}=G(\boldsymbol{x})\boldsymbol{\theta}+f(\boldsymbol{x})$ will allow us to prune regions in the parameter where $\boldsymbol{\theta}$ belongs.
    \item \textbf{Non-negativity of Lagrange multipliers:} \label{Non-negativity} Likewise from dual feasibility \eqref{dual_feasibility1}, recall that ${\mu^*_{j{\boldsymbol{\theta}}}} \geq 0$ $\forall j \in \{1,2,\cdots,M\}$. Since ${\mu^*_{j{\boldsymbol{\theta}}}}$ depends on $\boldsymbol{\theta}$, deriving an explicit expression for $\mu^*_{j{\boldsymbol{\theta}}}$ as a function of $\boldsymbol{\theta}$ will allow us to further prune the region where $\boldsymbol{\theta}$ belongs.
\end{enumerate}
Thus we define the \textit{instantaneous feasible set} $\boldsymbol{\Omega}(t)$ as
\begin{align}
\label{omegaspecial}
    \boldsymbol{\Omega}(t) \coloneqq \{\boldsymbol{\hat{\theta}} \in \mathbb{R}^p \vert  \boldsymbol{a}^T_{j}(t)\boldsymbol{u}^*_{\boldsymbol{\theta}}(t)-b_{j}(t)<0\forall j\in \mathcal{IA}(\boldsymbol{u}^*_{\boldsymbol{\theta}}(t)),-\mu^*_{j{\boldsymbol{\theta}}}(t)\leq 0 \forall j \in \{1,\cdots,M\}\}
\end{align}
\subsection{Using KKT conditions for deriving  $\boldsymbol{u}^*_{\boldsymbol{\theta}}$ and ${\mu^*_{j{\boldsymbol{\theta}}}}$}
The following discussion summarizes how we can simplify problem \eqref{optimization_formulation_1_static_obstacles} to a simpler problem so as to derive explicit expressions for both $\boldsymbol{u}^*_{\boldsymbol{\theta}}$ and ${\mu^*_{j{\boldsymbol{\theta}}}}$.  Let there be a total of $K$ active constraints \textit{i.e.}  $\texttt{card}(\mathcal{A}(\boldsymbol{u}^*_{\boldsymbol{\theta}}))=K$ where $K\in \{0,1,\cdots,M\}$. Using \eqref{dual_feasibility1} and \eqref{complimentarty slackness1}, we deduce
\begin{align}
	\mu^*_{j\boldsymbol{\theta}} = 0 \mbox{ $\forall j $} \in \mathcal{IA}(\boldsymbol{u}^*_{\boldsymbol{\theta}}).
\end{align}  
Therefore, we can restrict the summation in \eqref{stationarity1} only to the set of active constraints \textit{i.e.}
\begin{align}
\label{kkt_general}
	\boldsymbol{u}^*_{\boldsymbol{\theta}} &= \hat{\boldsymbol{u}} - \frac{1}{2}\sum_{j \in \mathcal{A}(\boldsymbol{u}^*_{\boldsymbol{\theta}})}\mu^*_{j{\boldsymbol{\theta}}}\boldsymbol{a}_{j}
	\nonumber \\
	&= \hat{\boldsymbol{u}} - \frac{1}{2} A_{ac}^{T} \boldsymbol{\mu}^{ac}_{\boldsymbol{\theta}} .
\end{align}
where $A_{ac}(\boldsymbol{x})\in \mathbb{R}^{K \times 2}$ is the matrix formed using the rows of $A$ that are indexed by the active set $\mathcal{A}(\boldsymbol{u}^*_{\boldsymbol{\theta}})$, and likewise $\boldsymbol{\mu}^{ac}_{\boldsymbol{\theta}}\coloneqq\{\mu^*_{j\boldsymbol{\theta}}\}_{j \in \mathcal{A}(\boldsymbol{u}^*_{\boldsymbol{\theta}})}$. Similarly, let $\boldsymbol{b}_{ac}(\boldsymbol{x})\in \mathbb{R}^{K}$ denote the vector formed from the elements of $\boldsymbol{b}$ indexed by $\mathcal{A}(\boldsymbol{u}^*_{\boldsymbol{\theta}})$. Likewise, we can define $A_{inac}(\boldsymbol{x})$ and $\boldsymbol{b}_{inac}(\boldsymbol{x})$ corresponding to the inactive set.  By deleting all inactive constraints and retaining only the active constraints from \eqref{optimization_formulation_1_static_obstacles}, we can pose another QP that consists only of active constraints, whose solution is the same as that of \eqref{optimization_formulation_1_static_obstacles}. This equality-constrained program (EQP) is given by
\begin{align}
\label{optimization_formulation_2_static_obstacles}
	\begin{aligned}
		\boldsymbol{u}^*&= \underset{\boldsymbol{u}}{\arg\min}
		& & \norm{\boldsymbol{u} - \hat{\boldsymbol{u}}(\boldsymbol{x})}^2 \\
		& \text{subject to}
		& & A_{ac}(\boldsymbol{x})\boldsymbol{u} = \boldsymbol{b}_{ac}(\boldsymbol{x})  
	\end{aligned}
\end{align}
The system $A_{ac}(\boldsymbol{x})\boldsymbol{u} = \boldsymbol{b}_{ac}(\boldsymbol{x})$ is always consistent by construction \eqref{activeinactive}, as long as a solution $\boldsymbol{u}^*_{\boldsymbol{\theta}}$ to \eqref{optimization_formulation_1_static_obstacles} exists. 
This EQP is useful because it is easier to derive an expression  $\boldsymbol{u}^*_{\boldsymbol{\theta}}(\boldsymbol{x})=G(\boldsymbol{x})\boldsymbol{\theta}+f(\boldsymbol{x})$ for  \eqref{optimization_formulation_2_static_obstacles} than the inequality constrained problem \eqref{optimization_formulation_1_static_obstacles}. The question is how can the observer estimate the active set $\mathcal{A}(\boldsymbol{u}^*_{\boldsymbol{\theta}})$ to determine $A_{ac}(\boldsymbol{x}),b_{ac}(\boldsymbol{x})$ for \eqref{optimization_formulation_2_static_obstacles}. This can be done as follows. From Assumption \ref{ass3}, recall that the observer can measure both the position $\boldsymbol{x}(t)$ and velocity $\boldsymbol{u}^*_{\boldsymbol{\theta}}(\boldsymbol{x}(t))$ of the robot. Using these, the observer can determine $\mathcal{A}(\boldsymbol{u}^*_{\boldsymbol{\theta}})$ by comparing residuals $\vert\boldsymbol{a}^T_{j}(\boldsymbol{x})\boldsymbol{u}^*_{\boldsymbol{\theta}} - b_{j}(\boldsymbol{x})\vert$ against a small threshold $\epsilon>0$ consistent with \eqref{activeinactive}:
\begin{align}
\label{activeobserver}
	\mathcal{A}^{obs.} \coloneqq \{j \in \{1,\cdots,M\} \mid \vert\boldsymbol{a}^T_{j}(\boldsymbol{x})\boldsymbol{u}^*_{\boldsymbol{\theta}} - b_{j}(\boldsymbol{x})\vert < \epsilon \} 
\end{align}
For a small threshold $\epsilon$, it holds true that $\mathcal{A}^{obs.}=	\mathcal{A}(\boldsymbol{u}^*_{\boldsymbol{\theta}})$ consistent with \eqref{activeinactive}. This allows the observer to determine the active set.  Next, we work with \eqref{optimization_formulation_2_static_obstacles} to derive an explicit expression for control \textit{i.e.} $\boldsymbol{u}^*_{\boldsymbol{\theta}}(\boldsymbol{x})=G(\boldsymbol{x})\boldsymbol{\theta}+f(\boldsymbol{x})$ and $\mu^*_{j{\boldsymbol{\theta}}}$ for various combinations of $\texttt{card}(\mathcal{A}(\boldsymbol{u}^*_{\boldsymbol{\theta}}))=K$ and $\texttt{rank}(A_{ac}(\boldsymbol{x}))$. 
\begin{table*}
	\caption{Enumerating cases where either point-wise identification or region-based identification or both are possible} \label{tab:sometab}
	\centering
	\begin{tabular}{|p{0.12\columnwidth}|p{0.22\columnwidth}|p{0.12\columnwidth}|p{0.2\columnwidth}|p{0.2\columnwidth}|}
		\hline
		\textbf{Case} & \textbf{Number of active constraints} & \textbf{\texttt{rank}$(A_{ac})$} & \textbf{Point-wise identification} & \textbf{Region-based identification} \   \\
		\hline
		A & None & - & Possible & \cellcolor{green!35} Possible \\
		\hline
		B & 1 & 1 & Possible & \cellcolor{green!35} Possible \\
		\hline
		C & $\geq 2$ & 1 & Possible & \cellcolor{green!35} Possible \\
		\hline
		D & $2$ & 2 & \cellcolor{red!35} Not possible & \cellcolor{green!35} Possible \\
		\hline
		E & $\geq 3$ & 2 & \cellcolor{red!35} Not possible & \cellcolor{green!35} Possible \\
		\hline
	\end{tabular}
\end{table*}
\section{Computing $\boldsymbol{\Omega}(t)$ using SVD of $A_{ac}(\boldsymbol{x})\boldsymbol{u} = \boldsymbol{b}_{ac}(\boldsymbol{x})$}
\label{SVD}
The aim of this section is to derive explicit expressions for the optimal control $\boldsymbol{u}^*_{\boldsymbol{\theta}}$ and Lagrange multipliers $\mu^*_{j{\boldsymbol{\theta}}}$ for region based estimation of $\boldsymbol{\theta}$. To derive these, we will bank on the EQP proposed in \eqref{optimization_formulation_2_static_obstacles} as a surrogate to the original problem in \eqref{optimization_formulation_1_static_obstacles}. We can further simplify the analysis of \eqref{optimization_formulation_2_static_obstacles} by considering different cases that can arise based on the number of active constraints ($K$) and linear independence relations among the constraints governed by $\texttt{rank}(A_{ac}(\boldsymbol{x}))$. Table \ref{tab:sometab} summarizes these cases along with whether point-wise estimation and feasible region-based estimation is possible when either of these arise in practice. Note that the last two rows where the number of active constraints are $\geq 2$ correspond to cases where region-based identification is the only method to infer bounds on the $\boldsymbol{\theta}$. As we will show, these in fact correspond to situations that are most likely to arise in multirobot systems applications as well, highlighting the importance of our proposed technique.

\subsection{No active constraints \textit{i.e.} $K=0$}
\label{caseA}
When no constraint is active, we have $\mu_{j\boldsymbol{\theta}}=0 \mbox{ } \forall j \in \{1,2,\cdots,M\}$. This means that  $\mathcal{A}(\boldsymbol{u}^*_{\boldsymbol{\theta}})=\phi$ and $\mathcal{IA}(\boldsymbol{u}^*_{\boldsymbol{\theta}})=\{1,2,\cdots,M\}$. In other words, the ego robot does not ``worry" about collisions with any other robot. From \eqref{stationarity1} we get $\boldsymbol{u}^*_{\boldsymbol{\theta}}=\hat{\boldsymbol{u}}(\boldsymbol{x})=C(\boldsymbol{x})\boldsymbol{\theta} + \boldsymbol{d}(\boldsymbol{x})$. From this expression, it is evident that the control $\boldsymbol{u}^*_{\boldsymbol{\theta}}$ and the robot dynamics $\boldsymbol{\dot{x}}$, exhibit a well-defined dependence on $\boldsymbol{\theta}$. Therefore, it \textit{is} possible to do point-wise identification of $\boldsymbol{\theta}$ using RLS, UKF or AO type of estimators (provided that $C(\boldsymbol{x})$ is persistently exciting \cite{narendra2012stable}). However, since point-wise identification is not under the purview of this paper, we focus on estimating regions where $\boldsymbol{\theta}$ can lie. Referring to \ref{KKTFeasible}, note that the condition for nonnegativity of Lagrange multipliers is satisfied trivially because $\mu_{j\boldsymbol{\theta}}=0 \mbox{ } \forall j \in \{1,2,\cdots,M\}$, so they do not give any information about $\boldsymbol{\theta}$. However, the condition for inactive constraints gives $\forall j \in \mathcal{IA}(\boldsymbol{u}^*_{\boldsymbol{\theta}})= \{1,2,\cdots,M\}$
\begin{align}
    \boldsymbol{a}^T_{j}\boldsymbol{u}^*_{\boldsymbol{\theta}}-b_{j} &<0   \nonumber \\
    \implies \boldsymbol{a}^T_{j}(C\boldsymbol{\theta} + \boldsymbol{d})-b_{j}&<0 \ \nonumber \\
    \implies (\boldsymbol{a}^T_{j}C)\boldsymbol{\theta} &< b_{j}-\boldsymbol{a}^T_{j}\boldsymbol{d}
    \label{caseAineq}
\end{align}
which represents a set of $M$ halfspace constraints on $\boldsymbol{\theta}$. Recall from \eqref{Ab_static} that $\boldsymbol{a}_{j}$ and $\boldsymbol{b}_{j}$ depend on the position of the ego robot \textit{i.e.} $\boldsymbol{x}$ and the positions of other robots \textit{i.e.} $\boldsymbol{x}^o_j$ $\forall j \in \{1,2,\cdots,M\} $. In \eqref{caseAineq}, we have omitted these dependencies to keep the notation light.
Thus, the desired set $\boldsymbol{\Omega}$ is defined using \eqref{Ab_static} and \eqref{omegaspecial} as follows
\begin{align}
\label{omegaspecial_CaseA}
    \boldsymbol{\Omega} \coloneqq \{\boldsymbol{\hat{\theta}} \in \mathbb{R}^p &\vert  (AC)\boldsymbol{\hat{\theta}} < \boldsymbol{b}-A\boldsymbol{d} \}
\end{align}
\subsection{Exactly one active constraint \textit{i.e.} $K=1$}
\label{caseB}
When exactly constraint is active \textit{i.e.} $K=1$, there is one obstacle that the ego robot ``worries" about for collision. Since there are two degrees of freedom in the control input, and one obstacle to avoid, the ego robot can avoid this obstacle and additionally minimize $\norm{\boldsymbol{u}-\hat{\boldsymbol{u}}(\boldsymbol{x})}^2$ with the remaining degree of freedom.  This causes $\boldsymbol{u}^*_{\boldsymbol{\theta}}$ to exhibit a well-defined dependence on $\hat{\boldsymbol{u}}(\boldsymbol{x})$ and by extension, on $\boldsymbol{\theta}$. This allows for point-wise identification of $\boldsymbol{\theta}$.

Region-based identification is also possible and can be used to expedite the convergence of point-wise identification. To derive the feasible region where $\boldsymbol{\theta}$ belongs, we need the expression for control $\boldsymbol{u}^*_{\boldsymbol{\theta}}$ and the Lagrange multiplier corresponding to the active constraint. Let $i \in \{1,2,\cdots,M\}$ denote the index of the active constraint. Thus, from \eqref{activeinactive}, we have $A_{ac}(\boldsymbol{x})\boldsymbol{u}^*_{\boldsymbol{\theta}}=\boldsymbol{a}^T_i(\boldsymbol{x})\boldsymbol{u}^*_{\boldsymbol{\theta}}=b_i(\boldsymbol{x})$  where $A_{ac}(\boldsymbol{x})\coloneqq \boldsymbol{a}^T_i(\boldsymbol{x}) $  and $\boldsymbol{a}^T_i(\boldsymbol{x}),
\boldsymbol{b}_i(\boldsymbol{x})$ are defined in  \eqref{Ab_static}. We use the SVD of $A_{ac}=U\Sigma V^T$.  Defining
\begin{align}
U &= 1 \nonumber \\
\Sigma &= \left[\begin{matrix} \Sigma_m,0 \end{matrix}\right] \mbox{ where }\Sigma_m=\norm{\boldsymbol{a}_i} \nonumber \\
V &=  \left[\begin{matrix} V_1,V_2 \end{matrix}\right] \mbox{ } V_1=\frac{\boldsymbol{a}_i}{\norm{\boldsymbol{a}_i}},V_2=R_{\frac{\pi}{2}}\frac{\boldsymbol{a}_i}{\norm{\boldsymbol{a}_i}}, 
\end{align}
Since $V$ forms a basis for $\mathbb{R}^2$, any $\boldsymbol{u}$ can be expressed as
\begin{align}
\boldsymbol{u}&=\left[\begin{matrix} V_1,V_2 \end{matrix}\right] \left[\begin{matrix} \tilde{u}_1 \\ \tilde{u}_2\end{matrix}\right]  \nonumber.  \\
\implies \boldsymbol{a}^T_i\boldsymbol{u}-b_i &=U\left[\begin{matrix} \Sigma_m,0 \end{matrix}\right] \left[\begin{matrix} V_1^T \\ V_2^T \end{matrix}\right] \left[\begin{matrix} V_1,V_2 \end{matrix}\right] \left[\begin{matrix} \tilde{u}_1 \\ \tilde{u}_2\end{matrix}\right]-b_i  \nonumber \\
&=U\Sigma_m\tilde{u}_1+ 0\cdot\tilde{u}_2-b_i=0
\end{align}
Choosing $\tilde{u}_1 = \Sigma_m^{-1}U^Tb_i$ and $\tilde{u}_2 = \psi \in \mathbb{R}$, we find that  
\begin{align}
\label{control_case2_general}
\boldsymbol{u}= V_1\Sigma_m^{-1}U^Tb_i + V_2\psi
\end{align}
satisfies $\boldsymbol{a}^T_i(\boldsymbol{x})\boldsymbol{u}=b_i(\boldsymbol{x})$ $ \forall \psi \in \mathbb{R}$. Recall from the properties of SVD that $V_2$ forms a basis for $\mathcal{N}(\boldsymbol{a}^T_i(\boldsymbol{x}))$. We tune $\psi$ to minimize $\norm{\boldsymbol{u} - \hat{\boldsymbol{u}}}^2$ by solving the following unconstrained minimization problem 
\begin{align}
\label{psi_determine}
\begin{aligned}
\psi^*&= \underset{\psi}{\arg\min}
& &\norm{\boldsymbol{u} - \hat{\boldsymbol{u}}}^2 \\
&= \underset{\psi}{\arg\min}  
& &\norm{V_1\Sigma_m^{-1}U^Tb_i + V_2\psi -\hat{\boldsymbol{u}} }^2,
\end{aligned}
\end{align}
which gives $\psi^*=V_2^T \hat{\boldsymbol{u}}$. Substituting this in \eqref{control_case2_general}, gives
\begin{align}
\label{control_case2_optimal}
\boldsymbol{u}^*_{\boldsymbol{\theta}} = V_1\Sigma_m^{-1}U^Tb_i + V_2V_2^T\hat{\boldsymbol{u}}.
\end{align}
This equation is the solution to \eqref{optimization_formulation_2_static_obstacles} and by extension, to \eqref{optimization_formulation_1_static_obstacles}. Substituting $\hat{\boldsymbol{u}}(\boldsymbol{x})=C(\boldsymbol{x})\boldsymbol{\theta} + \boldsymbol{d}(\boldsymbol{x})$ in \eqref{control_case2_optimal} gives
\begin{align}
\label{control_case2_optimalucap}
\boldsymbol{u}^*_{\boldsymbol{\theta}} &= V_1\Sigma_m^{-1}U^Tb_i + V_2V_2^T(C\boldsymbol{\theta} + \boldsymbol{d}) \nonumber \\
&=\underbrace{V_2V_2^TC}_{G} \boldsymbol{\theta} + \underbrace{V_1\Sigma_m^{-1}U^Tb_i+V_2V_2^T\boldsymbol{d}}_{\boldsymbol{f}} \nonumber \\
&=G\boldsymbol{\theta} + \boldsymbol{f}
\end{align}
From inactive constraints, we get $\forall j \in \mathcal{IA}(\boldsymbol{u}^*_{\boldsymbol{\theta}})$: 
\begin{align}
\label{inactive_caseB}
    \boldsymbol{a}^T_{j}\boldsymbol{u}^*_{\boldsymbol{\theta}}-b_{j} &<0   \nonumber \\
    \implies \boldsymbol{a}^T_{j}(G\boldsymbol{\theta} + \boldsymbol{f})-b_{j}&<0 \ \nonumber \\
    \implies (\boldsymbol{a}^T_{j}G)\boldsymbol{\theta} &< b_{j}-\boldsymbol{a}^T_{j}\boldsymbol{f}  \nonumber \\
    \implies (A_{inac}G)\boldsymbol{\theta} &\prec \boldsymbol{b}_{inac}-A_{inac}\boldsymbol{f} ,
\end{align}
where $A_{inac},\boldsymbol{b}_{inac}$ are the rows of $A,\boldsymbol{b}$ indexed by the inactive constraints $\mathcal{IA}(\boldsymbol{u}^*_{\boldsymbol{\theta}})$.  To get the Lagrange multiplier $\mu^*_{i\boldsymbol{\theta}}$, we use \eqref{kkt_general} and \eqref{control_case2_optimalucap},
\begin{align}
\label{LMcaseB}
	\boldsymbol{u}^*_{\boldsymbol{\theta}} 
	&= \hat{\boldsymbol{u}} - \frac{1}{2} A^T_{ac} {\mu}^{*}_{i\boldsymbol{\theta}}  \nonumber \\
	\implies \frac{1}{2} A^T_{ac} {\mu}^{*}_{i\boldsymbol{\theta}} &= \underbrace{(C-G)}_{\tilde{G}}\boldsymbol{\theta} + \underbrace{(\boldsymbol{d}-\boldsymbol{f})}_{\boldsymbol{\tilde{f}}} \nonumber \\
	&=\tilde{G}\boldsymbol{\theta} + \boldsymbol{\tilde{f}} \nonumber \\
	\implies {\mu}^{*}_{i\boldsymbol{\theta}} &=2\frac{A_{ac}(\tilde{G}\boldsymbol{\theta} + \boldsymbol{\tilde{f}})}{\norm{A^T_{ac}}^2}
\end{align}
From non-negativity of ${\mu}^{*}_{i\boldsymbol{\theta}}$ we get
\begin{align}
\label{LMnonnegcaseB}
    {\mu}^{*}_{i\boldsymbol{\theta}} \geq 0 \iff -A_{ac}\tilde{G} \boldsymbol{\theta} \leq A_{ac}\boldsymbol{\tilde{f}}
\end{align}
Thus the desired set $\boldsymbol{\Omega}$ where $\boldsymbol{\theta}$ belongs is defined using \eqref{inactive_caseB} and \eqref{LMnonnegcaseB} as follows
\begin{align}
\label{omegaspecialCaseB}
    \boldsymbol{\Omega} \coloneqq \{\boldsymbol{\hat{\theta}} \in \mathbb{R}^p \vert  (A_{inac}G)\boldsymbol{\hat{\theta}} \prec \boldsymbol{b}_{inac}-A_{inac}\boldsymbol{f}, -A_{ac}\tilde{G} \boldsymbol{\hat{\theta}} \leq A_{ac}\boldsymbol{\tilde{f}} \}
\end{align}
\subsection{$2\leq K\leq M$ and $\texttt{rank}(A_{ac}(\boldsymbol{x}))=1$}
\label{caseC}
Let's consider the more general case in which there is more than one constraint active, but all these constraints are linearly dependent on one constraint among them. That is to say there is effectively only one ``representative constraint". Consequently, this case is similar to the case with just one active obstacle. We now derive expressions for both $\boldsymbol{u}^*_{\boldsymbol{\theta}}$ and $\boldsymbol{\mu}^{ac}_{\boldsymbol{\theta}}$.  Let $i_1,i_2,\cdots,i_K\in\{1,\cdots,M\}$ be the indices of the $K$ active constraints which satisfy
\begin{align}
\label{abK}
\left[\begin{matrix}
\boldsymbol{a}^T_{i_1}(\boldsymbol{x}) 
\\ \boldsymbol{a}^T_{i_2}(\boldsymbol{x}) \\
\vdots \\
\boldsymbol{a}^T_{i_K}(\boldsymbol{x}) 
\end{matrix}\right]\boldsymbol{u}^*_{\boldsymbol{\theta}}&= 
\left[\begin{matrix}b_{i_1}(\boldsymbol{x}) \\b_{i_2}(\boldsymbol{x}) \\ \vdots \\ b_{i_K}(\boldsymbol{x}) \end{matrix}\right]  \mbox{ or } \nonumber \\
A_{ac}(\boldsymbol{x})\boldsymbol{u}^*_{\boldsymbol{\theta}}&=\boldsymbol{b}_{ac}(\boldsymbol{x}),
\end{align}
where $\boldsymbol{a}^T_{i_j}(\boldsymbol{x})  \mbox{ and } b_{i_j}(\boldsymbol{x})$ are defined using \eqref{Ab_static}. Using the SVD of $A_{ac}$ \eqref{abK}, we can derive an expression for $\boldsymbol{u}^*_{\boldsymbol{\theta}}(\boldsymbol{x})$. Let $A_{ac}=U\Sigma V^T$ where
\begin{align}
\label{svdCaseB}
U&\coloneqq  \left[\begin{matrix} U_1,U_2 \end{matrix}\right] \mbox{ where, } U_1\in \mathbb{R}^{K \times 1}, U_2\in \mathbb{R}^{K \times K-1} \nonumber \\
\Sigma &\coloneqq 
\left[\begin{array}{@{}c|c@{}}
\Sigma_r& 0^{1 \times 1} \\
\hline
0^{K-1 \times 1}  &0^{K-1 \times 1} 
\end{array}\right] \nonumber \\
V&\coloneqq  \left[\begin{matrix} V_1,V_2 \end{matrix}\right] \mbox{ where, } V_1,V_2 \in \mathbb{R}^{2}. 
\end{align}
Choosing $\boldsymbol{u}=V_1\tilde{u}_1+V_2\tilde{u}_2$, from \eqref{abK} we get
\begin{align}
A_{ac}\boldsymbol{u}-\boldsymbol{b}_{ac} &= \left[\begin{matrix} U_1,U_2 \end{matrix}\right]   \left[\begin{array}{@{}c|c@{}}
\Sigma_r & 0 \\
\hline
0  &0
\end{array}\right]  \left[\begin{matrix} V^T_1 \\ V^T_2 \end{matrix}\right] \left[\begin{matrix} V_1,V_2 \end{matrix}\right]  \left[\begin{matrix} \tilde{u}_1 \\ \tilde{u}_2 \end{matrix}\right] -\boldsymbol{b}_{ac} \nonumber \\
&= \left[\begin{matrix} U_1,U_2 \end{matrix}\right] \bigg(\left[\begin{matrix} \Sigma_r\tilde{u}_1 \\0 \end{matrix}\right]  -\left[\begin{matrix} U^T_1\boldsymbol{b}_{ac}\\U^T_2\boldsymbol{b}_{ac} \end{matrix}\right]\bigg).
\end{align}
Since $\norm{}^2$ is unitary invariant, from \eqref{abK}
\begin{align}
\norm{A_{ac}\boldsymbol{u}-\boldsymbol{b}_{ac} }^2 & = \norm{\left[\begin{matrix} \Sigma_r\tilde{u}_1 \\0 \end{matrix}\right]  -\left[\begin{matrix} U^T_1\boldsymbol{b}_{ac}\\U^T_2\boldsymbol{b}_{ac} \end{matrix}\right]}^2 \nonumber \\
&= \norm{\Sigma_r\tilde{u}_1 -U^T_1\boldsymbol{b}_{ac} }^2 + \norm{U^T_2\boldsymbol{b}_{ac}}^2 .
\end{align}
The minimum norm is achieved for $\tilde{u}_1=\Sigma_r^{-1}U_1^T\boldsymbol{b}_{ac}$. Choosing $\tilde{u}_2 =\psi \in \mathbb{R}$, the ``least-squares" solutions are
\begin{align}
\boldsymbol{u}&=V_1\Sigma_r^{-1}U_1^T\boldsymbol{b}_{ac} + V_2\psi.
\end{align}
Computing $\psi$ by minimizing $\norm{\boldsymbol{u} - \hat{\boldsymbol{u}}}^2$, we get $\psi^*=V_2^T\hat{\boldsymbol{u}}$ which gives 
\begin{align}
\label{control_case3_optimal}
\boldsymbol{u}^*_{\boldsymbol{\theta}}&=V_1\Sigma_r^{-1}U_1^T\boldsymbol{b}_{ac} + V_2V_2^T\hat{\boldsymbol{u}}.
\end{align}
This control is the solution to \eqref{optimization_formulation_2_static_obstacles} and by extension, to \eqref{optimization_formulation_1_static_obstacles}. Substituting $\hat{\boldsymbol{u}}(\boldsymbol{x})=C(\boldsymbol{x})\boldsymbol{\theta} + \boldsymbol{d}(\boldsymbol{x})$ in \eqref{control_case3_optimal} gives
\begin{align}
\label{control_case3_optimalucap}
\boldsymbol{u}^*_{\boldsymbol{\theta}} &= V_1\Sigma_r^{-1}U_1^T\boldsymbol{b}_{ac}+ V_2V_2^T(C\boldsymbol{\theta} + \boldsymbol{d}) \nonumber \\
&=\underbrace{V_2V_2^TC}_{G}\boldsymbol{\theta} + \underbrace{V_1\Sigma_r^{-1}U_1^T\boldsymbol{b}_{ac}+V_2V_2^T\boldsymbol{d}}_{\boldsymbol{f}} \nonumber \\
&=G\boldsymbol{\theta} + \boldsymbol{f}
\end{align}
Following the approach in \eqref{inactive_caseB}, for the inactive constraints, we get: 
\begin{align}
\label{inactive_caseC}
(A_{inac}G)\boldsymbol{\theta} &\prec \boldsymbol{b}_{inac}-A_{inac}\boldsymbol{f} 
\end{align}
To get the Lagrange multipliers $\boldsymbol{\mu}^{ac}_{\boldsymbol{\theta}} $, we use \eqref{kkt_general} and \eqref{control_case3_optimalucap},
\begin{align}
\label{LMcaseC}
	\boldsymbol{u}^*_{\boldsymbol{\theta}} 
	&= \hat{\boldsymbol{u}} - \frac{1}{2} A^T_{ac} \boldsymbol{\mu}^{ac}_{\boldsymbol{\theta}}  \nonumber \\
	\implies \frac{1}{2} A^T_{ac} \boldsymbol{\mu}^{ac}_{\boldsymbol{\theta}} &= (C-G)\boldsymbol{\theta} + (\boldsymbol{d}-\boldsymbol{f}) \nonumber \\
	&=\tilde{G}\boldsymbol{\theta} + \boldsymbol{\tilde{f}} \nonumber \\
	\implies A^T_{ac} \boldsymbol{\mu}^{ac}_{\boldsymbol{\theta}} &= 2(\tilde{G}\boldsymbol{\theta} + \boldsymbol{\tilde{f}}) \nonumber \\
	\implies A^T_{ac}\boldsymbol{\mu}^{ac}_{\boldsymbol{\theta}} &= \boldsymbol{w} 
\end{align}
We will call the right handside $2(\tilde{G}\boldsymbol{\theta} + \boldsymbol{\tilde{f}})=\boldsymbol{w}$ for convenience of notation. Here, given the fact that $A^T_{ac} \in \mathbb{R}^{2 \times K}$ and $\texttt{rank}(A^T_{ac})=1$, the Lagrange multipliers are underdetermined. We use the SVD of $A^T_{ac}=\tilde{U}\tilde{\Sigma} \tilde{V}^T$ to deduce an expression for $\boldsymbol{\mu}^{ac}_{\boldsymbol{\theta}}$. Here
\begin{align}
\label{svdCaseBAacT}
\tilde{U}&=  \left[\begin{matrix} \tilde{U}_1,\tilde{U}_2 \end{matrix}\right] \mbox{ where, } \tilde{U}_1, \tilde{U}_2\in \mathbb{R}^{2} \nonumber \\
\tilde{\Sigma} &= 
\left[\begin{array}{@{}c|c@{}}
\tilde{\Sigma}_r& 0^{1 \times K-1} \\
\hline
0^{1 \times 1}  &0^{1 \times K-1} 
\end{array}\right] \nonumber \\
\tilde{V}&=  \left[\begin{matrix} \tilde{V}_1,\tilde{V}_2 \end{matrix}\right] \mbox{ where, } \tilde{V}_1 \in \mathbb{R}^{K \times 1}, \tilde{V}_2 \in \mathbb{R}^{K \times K-1}.
\end{align}
As before we represent, $\boldsymbol{\mu}=\tilde{V}_1\tilde{\boldsymbol{\mu}_{1}}+\tilde{V}_2\tilde{\boldsymbol{\mu}_{2}}$.  Since $\norm{}^2$ is unitary invariant, from \eqref{LMcaseC}
\begin{align}
\norm{A^T_{ac}\boldsymbol{\mu}-\boldsymbol{w}}^2 & = \norm{\left[\begin{matrix} \Sigma_r\boldsymbol{\tilde{\mu}}_1 \\0 \end{matrix}\right]  -\left[\begin{matrix} \tilde{U}^T_1\boldsymbol{w}\\\tilde{U}^T_2\boldsymbol{w} \end{matrix}\right]}^2 \nonumber \\
&= \norm{\Sigma_r\boldsymbol{\tilde{\mu}}_1 -\tilde{U}^T_1\boldsymbol{w}}^2 + \norm{\tilde{U}^T_2\boldsymbol{w}}^2 .
\end{align}
The minimum norm is achieved for $\boldsymbol{\tilde{\mu}}_1=\Sigma_r^{-1}\tilde{U}_1^T\boldsymbol{w}$. Choosing $\boldsymbol{\tilde{\mu}}_2 =\boldsymbol{\eta} \in \mathbb{R}$, the ``least-squares" solutions are
\begin{align}
\boldsymbol{\mu}&=V_1\Sigma_r^{-1}\tilde{U}_1^T\boldsymbol{w} + V_2\boldsymbol{\eta}.
\end{align}
Then using \eqref{LMcaseC}, we get
\begin{align}
\boldsymbol{\mu}^{ac}_{\boldsymbol{\theta}} &= 2\tilde{V}_1\tilde{\Sigma}_r^{-1}\tilde{U}_1^T(\tilde{G}\boldsymbol{\theta}+\boldsymbol{\tilde{f}})+ \tilde{V}_2\boldsymbol{\eta},
\end{align}
where $\boldsymbol{\eta} \in \mathbb{R}^{K-1}$ are floating variables which can assume any value. 
From non-negativity of $\boldsymbol{\mu}^{ac}_{\boldsymbol{\theta}}$ we get
\begin{align}
\label{LMnonnegcaseC}
   \boldsymbol{\mu}^{ac}_{\boldsymbol{\theta}} \succeq 0 \iff -2\tilde{V}_1\tilde{\Sigma}_r^{-1}\tilde{U}_1^T\tilde{G}\boldsymbol{\theta} - \tilde{V}_2\boldsymbol{\eta}  \preceq 2\tilde{V}_1\tilde{\Sigma}_r^{-1}\tilde{U}_1^T\boldsymbol{\tilde{f}}
\end{align}
Thus, the desired set $\boldsymbol{\Omega}$ where $\boldsymbol{\theta}$ belongs is defined using \eqref{inactive_caseC} and \eqref{LMnonnegcaseC} as follows
\begin{align}
\label{omegaspecialCaseC}
    \boldsymbol{\Omega} \coloneqq \{\boldsymbol{\hat{\theta}} \in \mathbb{R}^p \vert  (A_{inac}G)\boldsymbol{\hat{\theta}} &< \boldsymbol{b}_{inac}-A_{inac}\boldsymbol{f},\nonumber \\ -2\tilde{V}_1\tilde{\Sigma}_r^{-1}\tilde{U}_1^T\tilde{G}\boldsymbol{\hat{\theta}} - \tilde{V}_2\boldsymbol{\eta}  &\preceq 2\tilde{V}_1\tilde{\Sigma}_r^{-1}\tilde{U}_1^T\boldsymbol{\tilde{f}} \mbox{ if $\exists$} \boldsymbol{\eta} \in \mathbb{R}^{K-1}  \}
\end{align}
\subsection{$K=2$ and $\texttt{rank}(A_{ac}(\boldsymbol{x}))=2$}
\label{caseD}
Consider the case where there are exactly \textit{two} constraints that are active and linearly independent. In this case, there are as many degrees of freedom in control as the number of independent active obstacles to avoid. Consequently, $ \boldsymbol{u}^*_{\boldsymbol{\theta}}$ is completely determined by the active constraints and hence does not depend on $\hat{\boldsymbol{u}}$.  We formally demonstrate this claim as follows.  Let $i_1,i_2 \in \{1,2,\cdots,M\}$ be the indices of the two constraints that are active. These constraints satisfy \eqref{abK} except that here $A_{ac}(\boldsymbol{x})\in \mathbb{R}^{2 \times 2}$ and $\texttt{rank}(A_{ac}(\boldsymbol{x}))=2$.  This problem is well-posed, its solution is given by 
\begin{align}
\label{control_caseD}
A_{ac}(\boldsymbol{x})\boldsymbol{u}^*_{\boldsymbol{\theta}} &= \boldsymbol{b}_{ac}(\boldsymbol{x}) \nonumber \\
\implies \boldsymbol{u}^*_{\boldsymbol{\theta}}&=A^{-1}_{ac}(\boldsymbol{x})\boldsymbol{b}_{ac}(\boldsymbol{x})
\end{align}
where the inverse exists because $\texttt{rank}(A_{ac}(\boldsymbol{x}))=2$. Since neither $A^{-1}_{ac}$ nor $\boldsymbol{b}_{ac}$ depend on $\boldsymbol{\theta}$ \eqref{Ab_static}, $\boldsymbol{u}^*_{\boldsymbol{\theta}}$ also does not depend on $\boldsymbol{\theta}$. Hence, point-wise identification is not possible. For feasible region-based identification, we cannot get information from inactive constraints because that is also contingent upon on $\boldsymbol{u}^*_{\boldsymbol{\theta}}$ depending on $\boldsymbol{\theta}$ (subsec. \ref{KKTFeasible}). Nevertheless, non-negativity of Lagrange multipliers is still useful. Since constraints $i_1$ and $i_2$ are active, it follows from dual feasibility \eqref{dual_feasibility1} and subsec. \ref{KKTFeasible} that $\mu^*_{i_1 \boldsymbol{\theta}} \geq 0$ and $\mu^*_{i_2 \boldsymbol{\theta}} \geq 0$. Define $\boldsymbol{\mu}^{ac}_{\boldsymbol{\theta}}=(\mu^*_{i_1 \boldsymbol{\theta}},\mu^*_{i_2 \boldsymbol{\theta}})$, then from \eqref{kkt_general} and \eqref{control_caseD}, we have
\begin{align}
    \boldsymbol{u}^*_{\boldsymbol{\theta}} &= \hat{\boldsymbol{u}} - \frac{1}{2} A_{ac}^{T} \boldsymbol{\mu}^{ac}_{\boldsymbol{\theta}} \nonumber \\
    \implies A_{ac}^{-1}\boldsymbol{b}_{ac}&=\hat{\boldsymbol{u}} - \frac{1}{2} A_{ac}^{T} \boldsymbol{\mu}^{ac}_{\boldsymbol{\theta}}\nonumber \\
    \implies \boldsymbol{\mu}^{ac}_{\boldsymbol{\theta}}&=2A_{ac}^{-T}(\hat{\boldsymbol{u}}-A_{ac}^{-1}\boldsymbol{b}_{ac}) \nonumber \\
    \implies \boldsymbol{\mu}^{ac}_{\boldsymbol{\theta}}&=2A_{ac}^{-T}(C\boldsymbol{\theta}+\boldsymbol{d}-A_{ac}^{-1}\boldsymbol{b}_{ac}) \nonumber\\
    \implies \boldsymbol{\mu}^{ac} \succeq \boldsymbol{0} &\iff -A_{ac}^{-T}C\boldsymbol{\theta} \preceq A_{ac}^{-T}(\boldsymbol{d}-A_{ac}^{-1}\boldsymbol{b}_{ac})
\end{align}
Thus the desired set $\boldsymbol{\Omega}$ is defined using \eqref{omegaspecial} as follows
\begin{align}
\label{omegaspecial_CaseD}
    \boldsymbol{\Omega} \coloneqq \{\boldsymbol{\hat{\theta}} \in \mathbb{R}^p &\vert  -A_{ac}^{-T}C\boldsymbol{\hat{\theta}} \preceq A_{ac}^{-T}(\boldsymbol{d}-A_{ac}^{-1}\boldsymbol{b}_{ac}) \}
\end{align}
where we have omitted the dependence on $\boldsymbol{x}$ and $\boldsymbol{x}^o_j$ to keep the notation light.
\subsection{$2< K\leq M$ and $\texttt{rank}(A_{ac}(\boldsymbol{x}))=2$}
\label{caseE}
Finally, let's consider the case where there are $K>2$ constraints that are active and two of them are linearly independent. The remaining active constraints are linear combinations of these two. In this case, like in case D, there are fewer degrees of freedom in control than the number of independent active obstacles to avoid. Consequently, neither $ \boldsymbol{u}^*_{\boldsymbol{\theta}}$ nor the robot dynamics depend on $\boldsymbol{\theta}$ making point-wise identification infeasible. Let $i_1,i_2,\cdots,i_K \in \{1,2,\cdots,M\}$ be the indices of the active constraints. These constraints satisfy \eqref{abK} except that here $\texttt{rank}(A_{ac}(\boldsymbol{x}))=2$.  This problem is well-posed albeit overdetermined, its solution is given by
\begin{align}
\label{control_caseE}
\boldsymbol{u}^*_{\boldsymbol{\theta}}&=A^{\dagger}_{ac}(\boldsymbol{x})\boldsymbol{b}_{ac}(\boldsymbol{x})
\end{align}
where $A^{\dagger}_{ac}$ denotes the Moore-Penrose pseudoinverse which exists because $\texttt{rank}(A_{ac}(\boldsymbol{x}))=2$. Since $\boldsymbol{u}^*_{\boldsymbol{\theta}}$ is independent of $\boldsymbol{\theta}$, point-wise identification is not possible. Likewise, for feasible region-based identification, we cannot get information from inactive constraints. Nevertheless, non-negativity of Lagrange multipliers is still useful. Since constraints $i_1,i_2,\cdots,i_K$ are active, it follows from dual feasibility \eqref{dual_feasibility1} and subsec. \ref{KKTFeasible} that $\mu^*_{i_1 \boldsymbol{\theta}},\cdots,\mu^*_{i_K \boldsymbol{\theta}} \geq 0$. From \eqref{kkt_general} and \eqref{control_caseE}, we have
\begin{align}
\label{LMcaseE}
    A_{ac}^{\dagger}\boldsymbol{b}_{ac}&=\hat{\boldsymbol{u}} - \frac{1}{2} A_{ac}^{T} \boldsymbol{\mu}^{ac}_{\boldsymbol{\theta}}\nonumber \\
    \implies  A_{ac}^{T} \boldsymbol{\mu}^{ac}_{\boldsymbol{\theta}} &= 2(C\boldsymbol{\theta}+\boldsymbol{d}-A_{ac}^{\dagger}\boldsymbol{b}_{ac}) \nonumber \\
    \implies  A_{ac}^{T} \boldsymbol{\mu}^{ac}_{\boldsymbol{\theta}} &= \boldsymbol{w}
\end{align}
We defined $\boldsymbol{w}=2(C\boldsymbol{\theta}+\boldsymbol{d}-A_{ac}^{\dagger}\boldsymbol{b}_{ac})$ to simplify notation. The above linear system for determining $\boldsymbol{\mu}^{ac}_{\boldsymbol{\theta}}$ is full rank but underdetermined because $\texttt{rank}(A^T_{ac})=2$. The SVD of $A^T_{ac}$ is $A^T_{ac}=\tilde{U}\tilde{\Sigma} \tilde{V}^T$ where
\begin{align}
\label{svdCaseE}
\tilde{U} &\in \mathbb{R}^{2 \times 2} \nonumber \\
\tilde{\Sigma} &= 
\left[\begin{array}{@{}c|c@{}}
\tilde{\Sigma}_m& 0^{2 \times K-2}
\end{array}\right] \mbox{ where, } \tilde{\Sigma}_m\in \mathbb{R}^{2 \times 2}\nonumber \\
\tilde{V}&=  \left[\begin{matrix} \tilde{V}_1,\tilde{V}_2 \end{matrix}\right] \mbox{ where, } \tilde{V}_1\in \mathbb{R}^{K \times 2},\tilde{V}_2 \in \mathbb{R}^{K \times K-2}. 
\end{align}
Since $\tilde{V}$ forms a basis for $\mathbb{R}^K$, any $\boldsymbol{\mu}$ can be expressed as
\begin{align}
\boldsymbol{\mu}&=\left[\begin{matrix} \tilde{V}_1,\tilde{V}_2 \end{matrix}\right] \left[\begin{matrix} \tilde{\boldsymbol{\mu}}_1  \\ \tilde{\boldsymbol{\mu}}_2 \end{matrix}\right]  =\tilde{V}_1\tilde{\boldsymbol{\mu}}_1 
+\tilde{V}_2\tilde{\boldsymbol{\mu}}_2 \nonumber.  \\
\implies A^{T}_{ac}\boldsymbol{\mu}-\boldsymbol{w} &=\tilde{U}\left[\begin{matrix} \Sigma_m,0 \end{matrix}\right] \left[\begin{matrix} \tilde{V}_1^T \\ \tilde{V}_2^T \end{matrix}\right] \left[\begin{matrix} \tilde{V}_1,\tilde{V}_2 \end{matrix}\right] \left[\begin{matrix} \tilde{\boldsymbol{\mu}}_1 \\ \tilde{\boldsymbol{\mu}}_2\end{matrix}\right]-\boldsymbol{w}  \nonumber \\
&=\tilde{U}\Sigma_m \tilde{\boldsymbol{\mu}}_1+ 0\cdot\tilde{\boldsymbol{\mu}}_2-\boldsymbol{w}=0
\end{align}
Choosing $\tilde{\boldsymbol{\mu}}_1 = \Sigma_m^{-1}\tilde{U}^T\boldsymbol{w}$ and $\tilde{\boldsymbol{\mu}}_2= \boldsymbol{\eta} \in \mathbb{R}^{K-2}$, we find that  
\begin{align}
\boldsymbol{\mu}^{ac}_{\boldsymbol{\theta}} &= \tilde{V}_1\tilde{\Sigma}_m^{-1}\tilde{U}^T\boldsymbol{w}+ \tilde{V}_2\boldsymbol{\eta} \nonumber \\
&= 2\tilde{V}_1\tilde{\Sigma}_m^{-1}\tilde{U}^T(C\boldsymbol{\theta}+\boldsymbol{d}-A_{ac}^{\dagger}\boldsymbol{b}_{ac})+ \tilde{V}_2\boldsymbol{\eta}
\end{align}
Here $\boldsymbol{\eta} \in \mathbb{R}^{K-2}$ are floating variables which can assume any value. 
From non-negativity of $\boldsymbol{\mu}^{ac}_{\boldsymbol{\theta}}$ we get
\begin{align}
\label{LMnonnegcaseE}
   &\boldsymbol{\mu}^{ac}_{\boldsymbol{\theta}} \succeq 0 \iff  -2\tilde{V}_1\tilde{\Sigma}_m^{-1}\tilde{U}^TC\boldsymbol{\theta}-\tilde{V}_2\boldsymbol{\eta}\preceq 2\tilde{V}_1\tilde{\Sigma}_m^{-1}\tilde{U}^T(\boldsymbol{d}-A_{ac}^{\dagger}\boldsymbol{b}_{ac}) 
\end{align}
Thus the desired set $\boldsymbol{\Omega}$ where $\boldsymbol{\theta}$ belongs is defined using \eqref{LMnonnegcaseE} as follows
\begin{align}
\label{omegaspecialCaseE}
    \boldsymbol{\Omega} \coloneqq \{\boldsymbol{\hat{\theta}} \in \mathbb{R}^p \vert -2\tilde{V}_1\tilde{\Sigma}_m^{-1}\tilde{U}^TC\boldsymbol{\hat{\theta}}-\tilde{V}_2\boldsymbol{\eta}\preceq 2\tilde{V}_1\tilde{\Sigma}_m^{-1}\tilde{U}^T(\boldsymbol{d}-A_{ac}^{\dagger}\boldsymbol{b}_{ac}) \mbox{ if $\exists$}\boldsymbol{\eta} \in \mathbb{R}^{K-2} \}
\end{align}
\section{Simulation Results}
\label{Results}

In this section, we present simulations to demonstrate the effectiveness of our proposed region-based parameter identification. We consider a multirobot system in which the task of each robot is to reach a goal position while avoiding collisions with other robots. The observer's problem is to infer the goal of each robot using the its positions and velocities as measurements. We use an Unscented Kalman Filter (UKF) as a baseline for goal estimation, to show how our region-based estimator outperforms a UKF.

To infer robot goals using the region-based estimator, the observer must estimate the instantaneous feasible region $\boldsymbol{\Omega}(t)$ using which it will compute the \textit{cumulative} feasible region $\boldsymbol{\Theta}(t)$. Recall from \eqref{timeintersection}, that $\boldsymbol{\Theta}(t)$ is computed by taking intersections of all $\boldsymbol{\Omega}(t)$ over time, and then its intersection with a compact set $\boldsymbol{\Theta}_0$ where the $\boldsymbol{\theta}$ is known to belong.  For the goal inference problem, the observer can compute the instantaneous set $\boldsymbol{\Omega}(t)$ using the results we derived in subsections \ref{caseA}-\ref{caseE} by checking how many obstacles are active at a given time \eqref{activeobserver}. As for $\boldsymbol{\Theta}_0$, we assume that the observer knows some reasonable upper and lower bounds on the location of the robot goals. In subsec. \ref{caseD}, we showed that when two or more obstacles are active at a given time, point-wise parameter identification is not possible because robot dynamics are independent of the underlying parameters. The feasible region based estimator, however, can give produce a bounded set where the underlying parameter belongs. To substantiate this, we first show results for a single robot navigating in an environment consisting of static obstacles. Subsequently, we will show results for multirobot inference as well.
\begin{figure*}
	\centering     
	\subfigure[$t=0.40s$]{\label{fig:a1}\includegraphics[trim=320 30 350 80, clip,width=.25\textwidth]{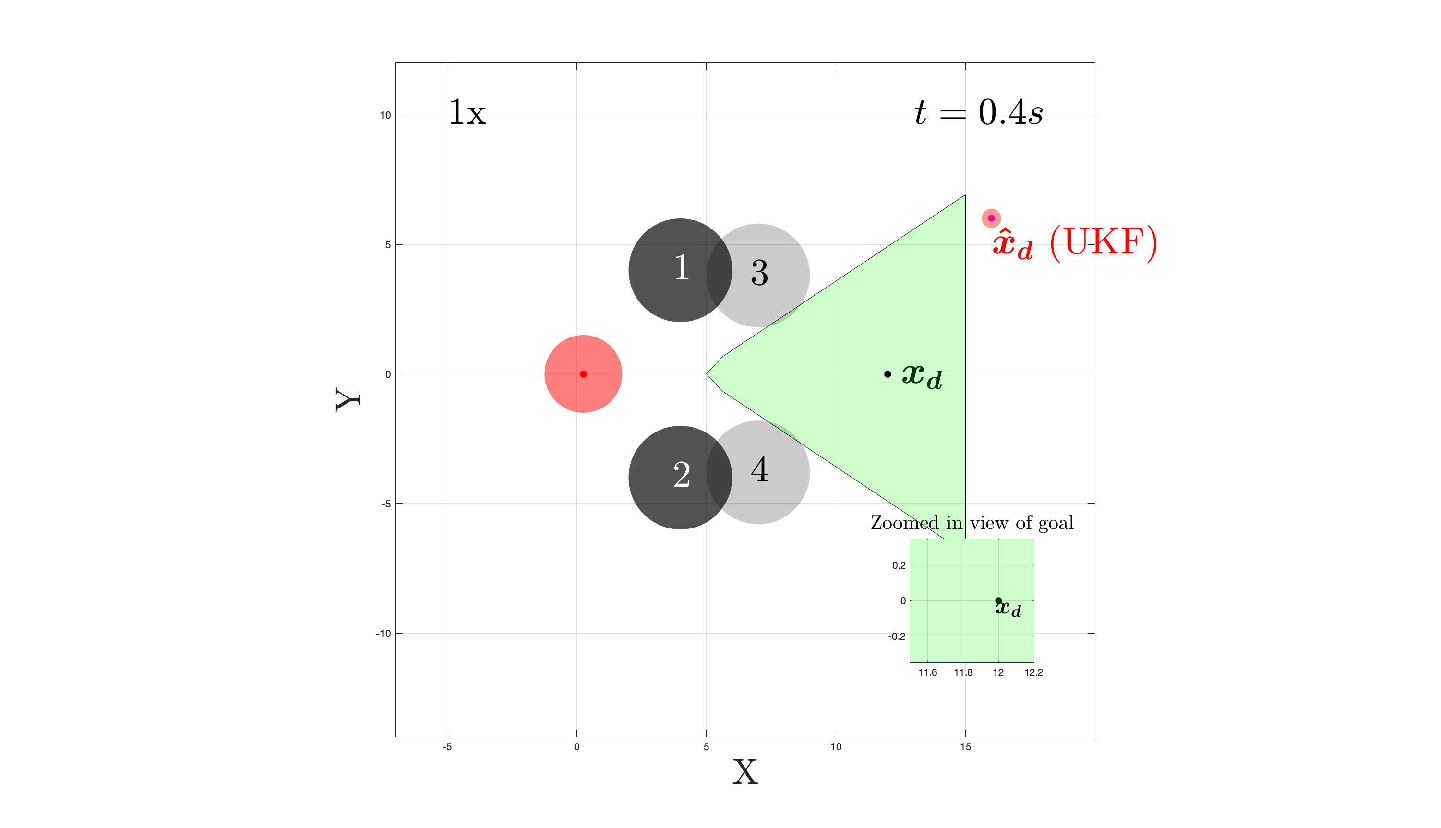}}
	\subfigure[$t=1.60s$]{\label{fig:b1}\includegraphics[trim=320 30 350 80, clip,width=.25\textwidth]{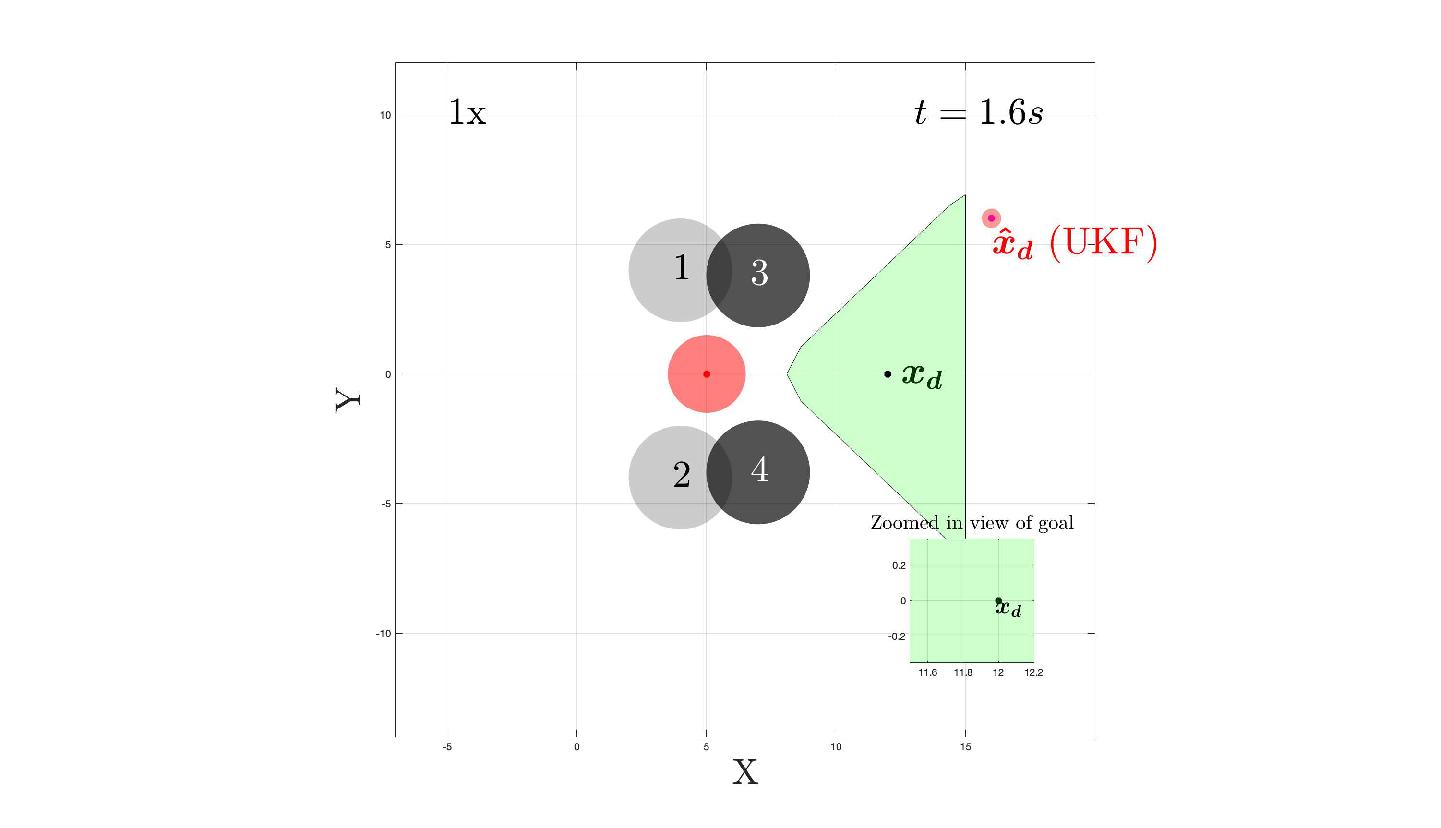}}
	\subfigure[$t=2.80s$]{\label{fig:c1}\includegraphics[trim=320 30 350 80, clip,width=.25\textwidth]{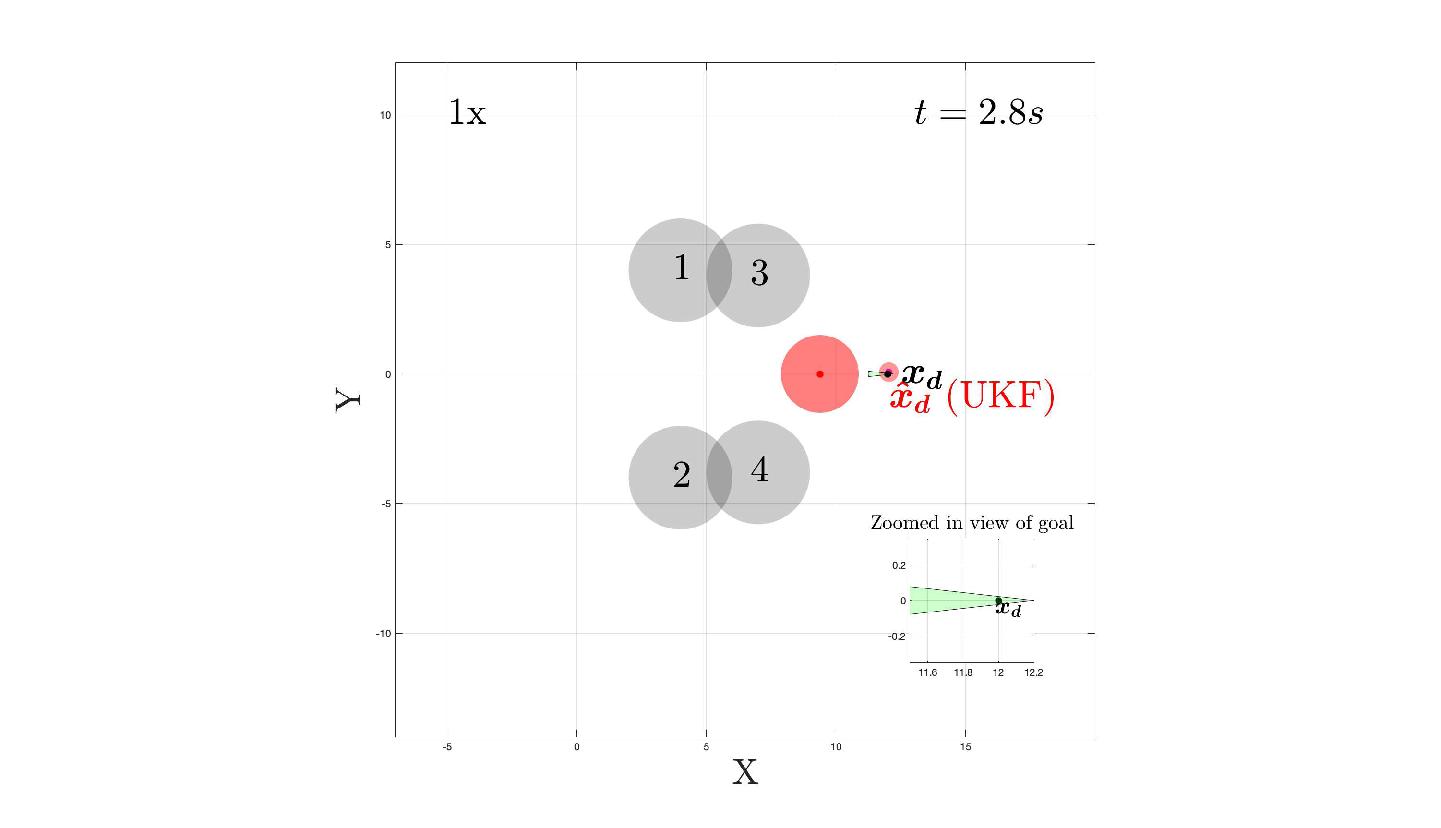}}
	\subfigure[Error Comparison]{\label{fig:d1}\includegraphics[width=0.23\textwidth,height=38mm]{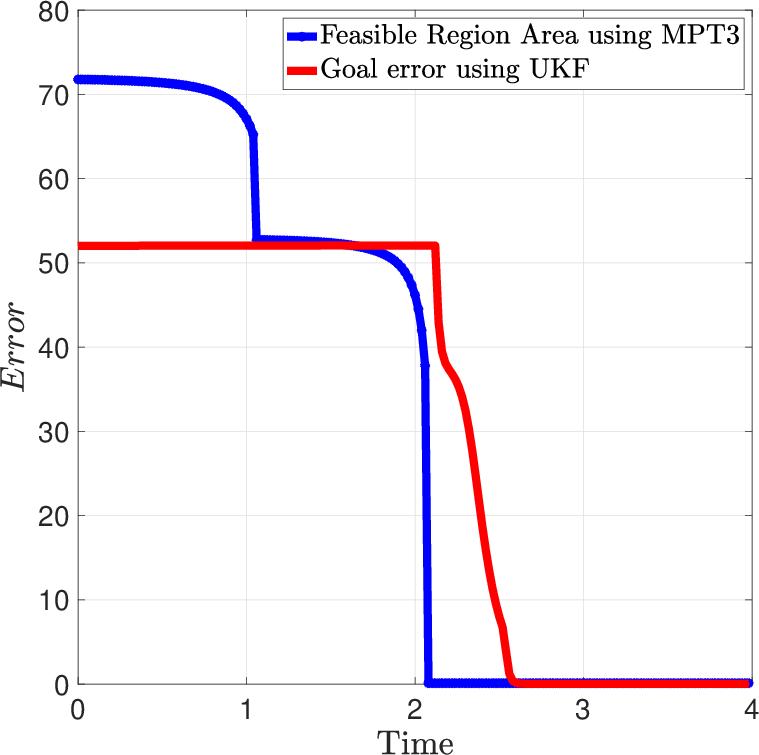}}
	\caption{(a)-(c) Goal identification for a robot navigating among static obstacles. Dark discs represent active obstacles. (d) Comparison of region-based estimation (blue) with a UKF.  Video at \url{https://youtu.be/jH3mxZhX2mA}}
	\label{twoactive}
\end{figure*}
\begin{figure*}
	\centering     
	\subfigure[$t=0.20s$]{\label{fig:a2}\includegraphics[trim=320 30 350 80, clip,width=.25\textwidth]{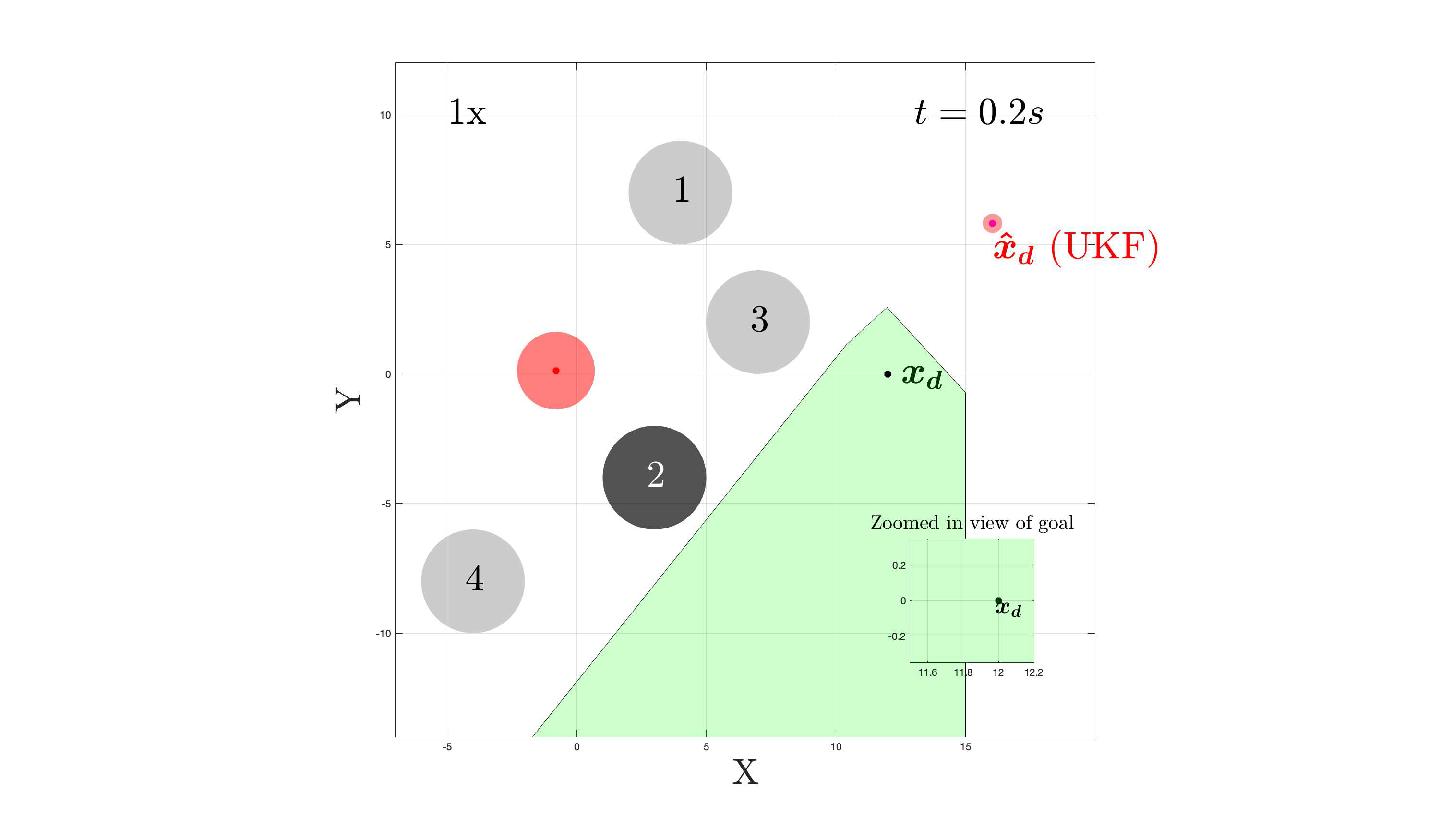}}
	\subfigure[$t=1.00s$]{\label{fig:b2}\includegraphics[trim=320 30 350 80, clip,width=.25\textwidth]{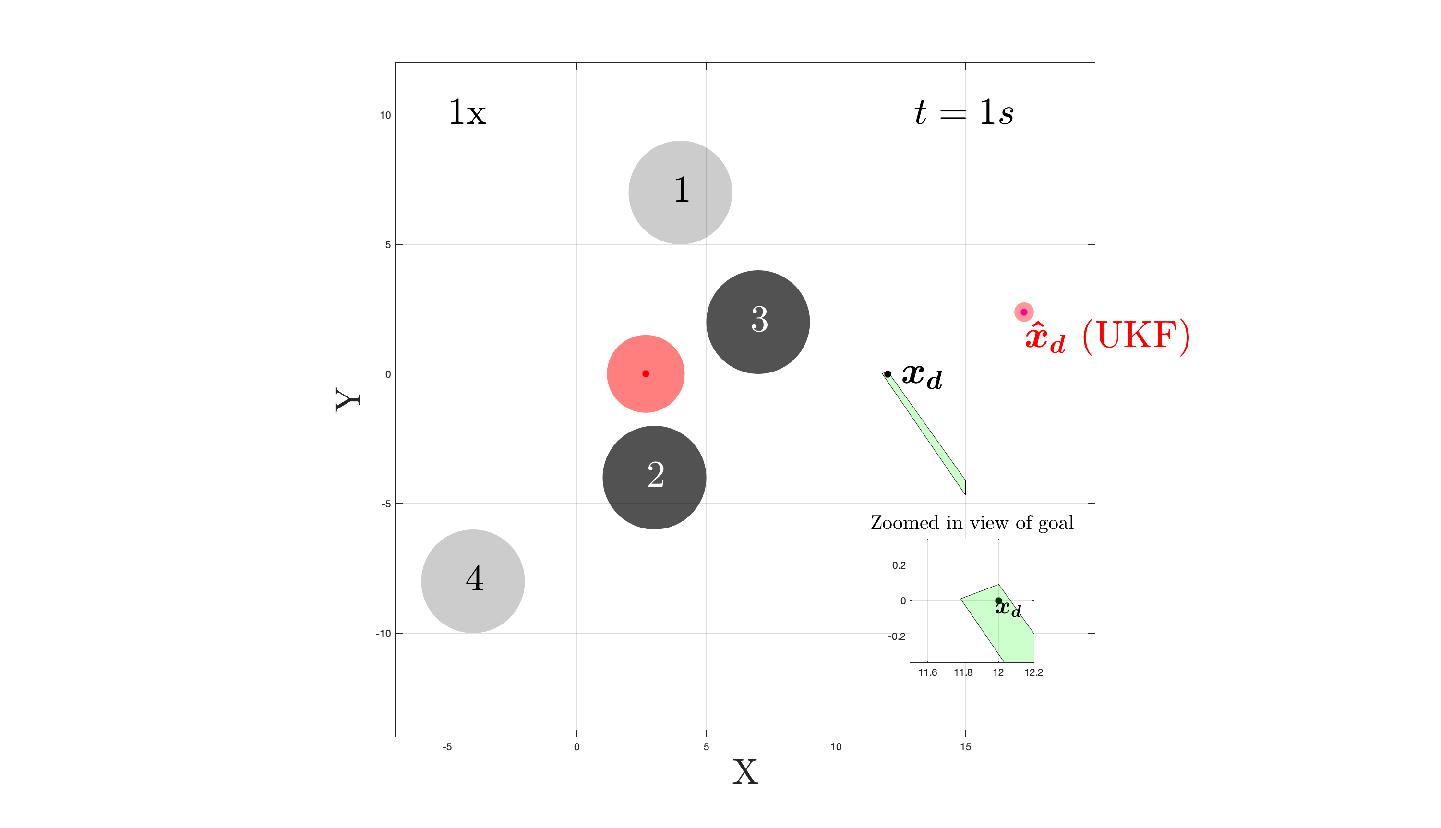}}
	\subfigure[$t=3.40s$]{\label{fig:c2}\includegraphics[trim=320 30 350 80, clip,width=.25\textwidth]{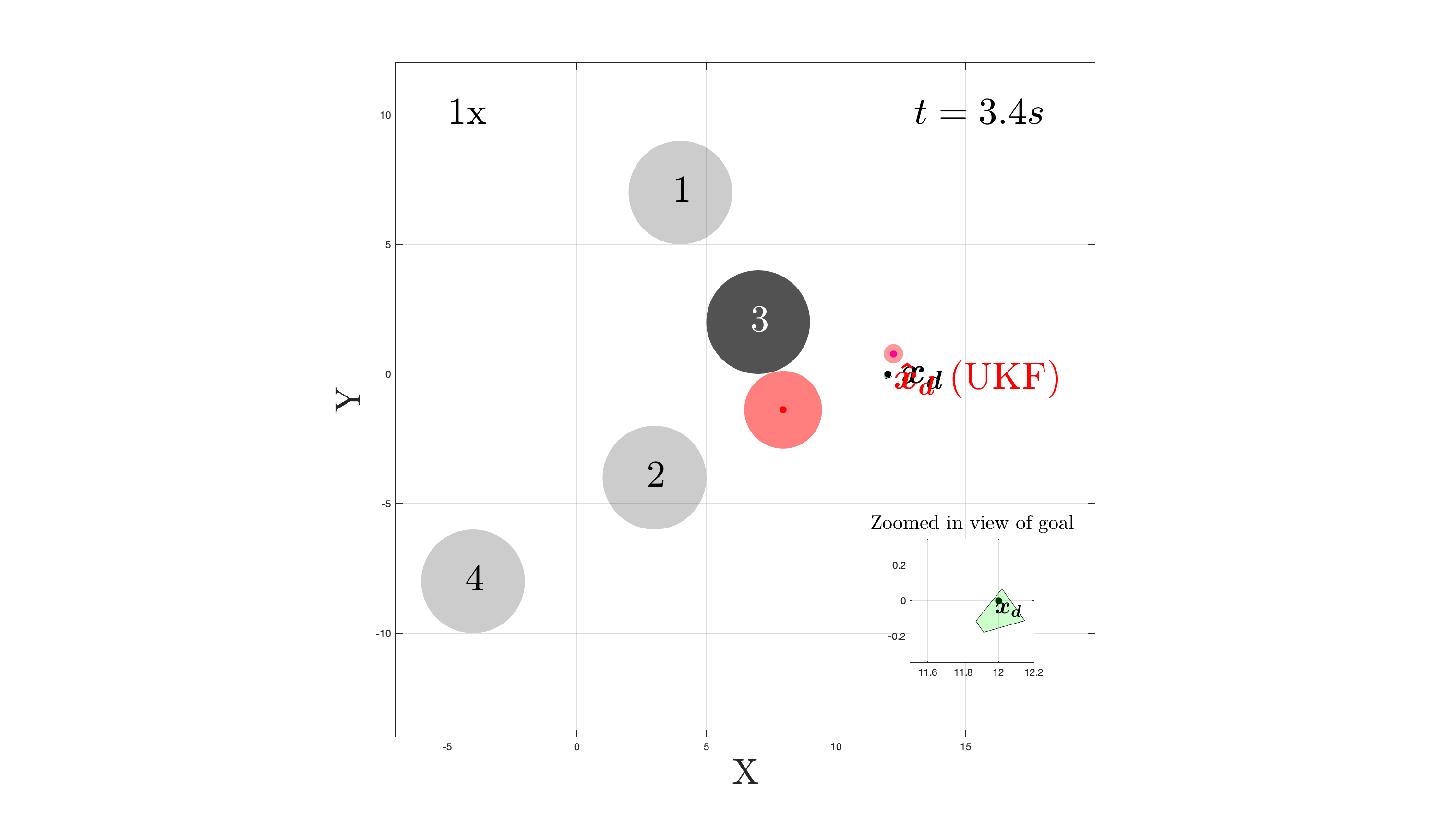}}
	\subfigure[Error Comparison]{\label{fig:d2}\includegraphics[width=0.23\textwidth,height=38mm]{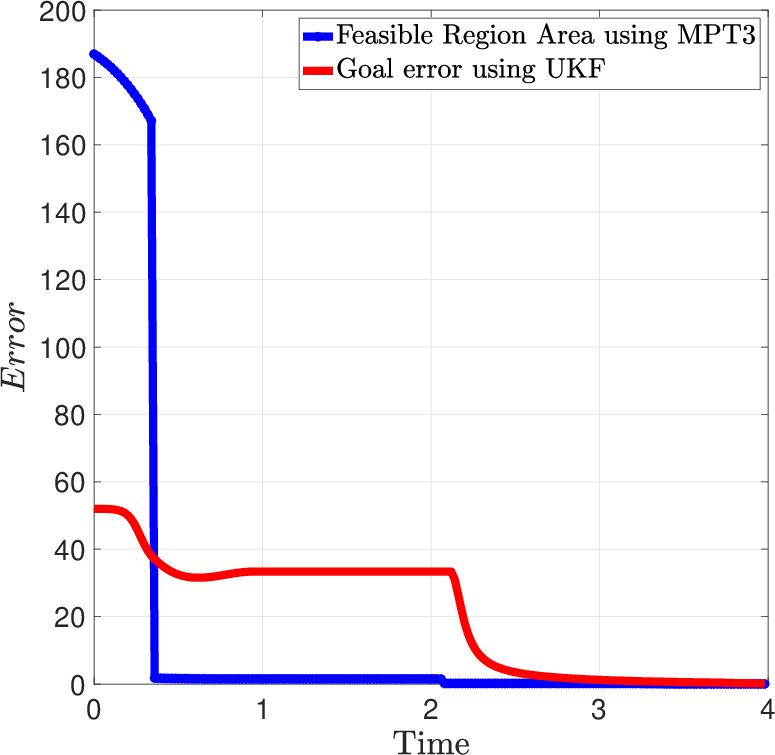}}
	\caption{(a)-(c) Goal identification for a robot navigating among static obstacles. Dark discs represent active obstacles. (d) Comparison of region-based estimation (blue) with a UKF (red). Video at \url{https://youtu.be/VthIXiBvjfU}}
	\label{mulactive}
\end{figure*}
In Figs. \ref{fig:a1}-\ref{fig:c1}, an ego robot (red) is trying to reach its goal $\boldsymbol{x}_d$ shown in black. The green regions correspond to $\boldsymbol{\Theta}(t)$ computed using the instantaneous feasible regions $\boldsymbol{\Omega}(t)$. As the robot moves to the right, obstacles one and two remain active until $t=1.1s$.  After this, obstacles three and four stay active until $t=2.1s$. Therefore, a point-wise estimator will not result in reduction in estimation error. Thus, until $t=2.1s$, the UKF based estimator (red) does not get updated as is evident from Figs. \ref{fig:a1}-\ref{fig:b1}. On the contrary, the green regions continue to shrink. The inset in these figures shows a zoomed in view around the goal to empirically demonstrate that the feasible region always remains non-empty and always encompasses the goal. In Fig. \ref{fig:c1}, all the obstacles are inactive so the UKF estimator will converge. Additionally, the feasible region based estimator will also provide feasible regions based on the set defined in subsec \ref{caseA}.   We measure the $\texttt{Vol}(\boldsymbol{\Theta}(t)$ of the feasible regions by computing the area of the green regions using the MPT3 toolbox in MATLAB. Fig. \ref{fig:d1} shows this area as a function of time and demonstrates the fast convergence as the robot moves, in comparison to the UKF estimator based error shown in red.

In Figs. \ref{fig:a2}-\ref{fig:c2}, we consider a different arrangement of obstacles. Here, until $t=0.38s$, only one obstacle is active; then until $t=2.08$s exactly two obstacles are active, then until $t=3.68s$ exactly one is active, following which all obstacles are inactive. As the robot moves, the region-based estimator converges quickly to the true goal as is evident in the insets in Figs. \ref{fig:a2}-\ref{fig:c2}. Further, the errors shown in \ref{fig:d2} illustrate how the UKF estimator takes some time to converge while the region based estimator converges much faster to the true goal. Furthermore, the UKF estimator does not respect the feasibility of the goal, it frequently remains outside the feasible region. This is to be expected because the derivation for UKF estimator does not consider non-negativity of Lagrange multipliers. \ref{fig:d2}.

\begin{figure}
	\centering     
	\subfigure[$t=0.08s$]{\label{fig:a3}\includegraphics[trim=320 30 350 80, clip,width=.24\textwidth]{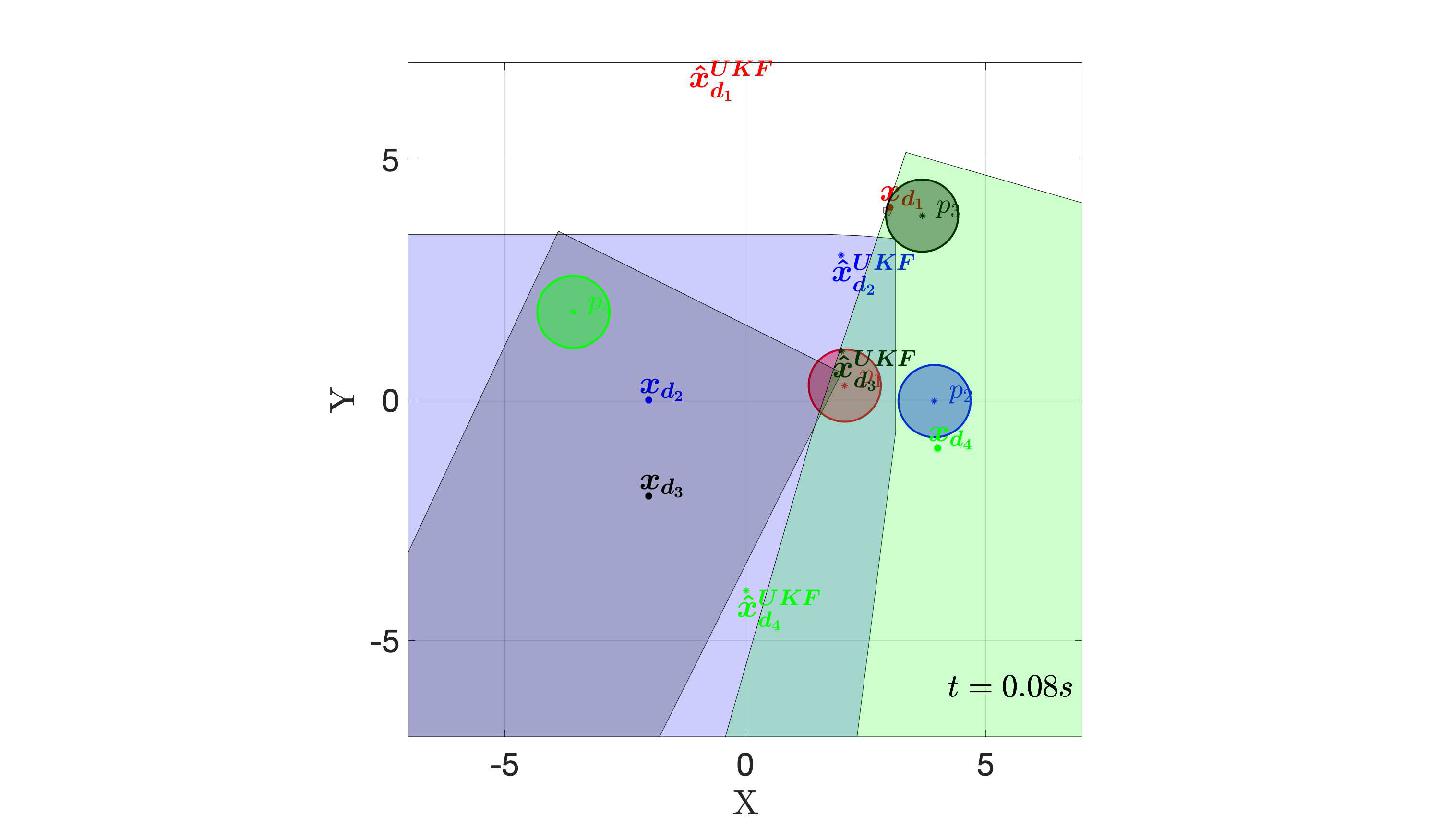}}
	\subfigure[$t=0.44s$]{\label{fig:b3}\includegraphics[trim=320 30 350 80, clip,width=.24\textwidth]{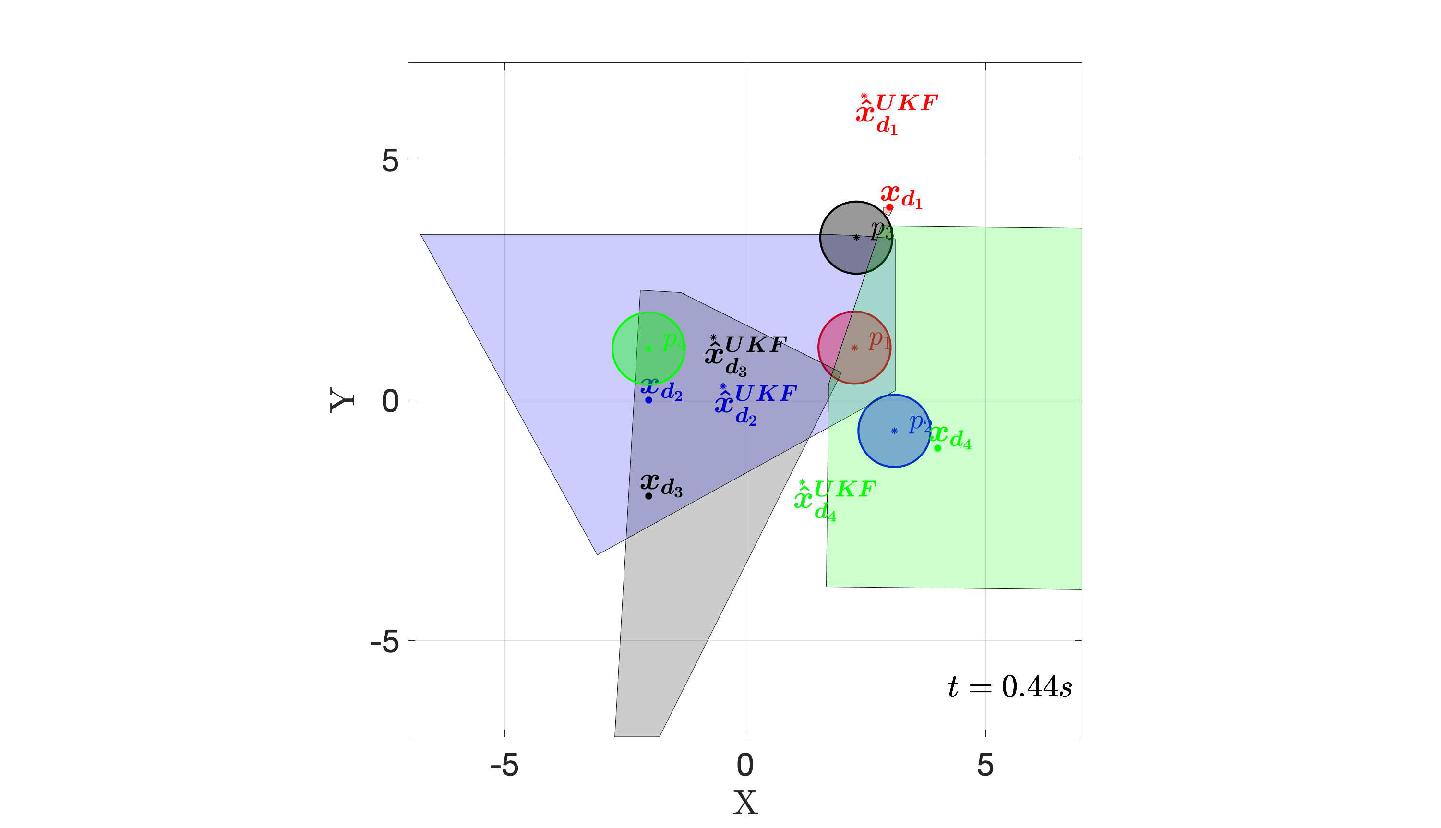}}
	\subfigure[$t=0.76s$]{\label{fig:c3}\includegraphics[trim=320 30 350 80, clip,width=.24\textwidth]{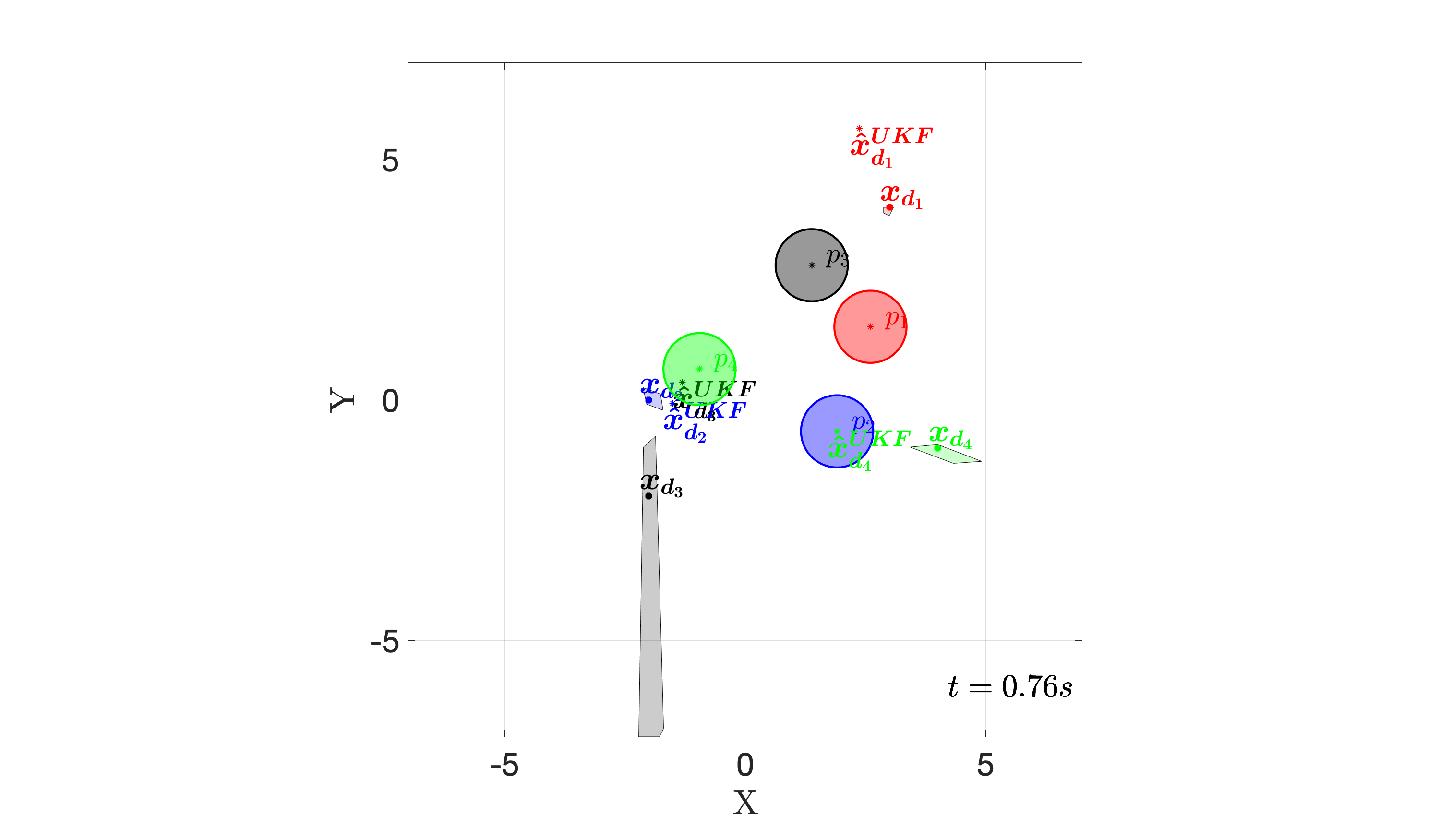}}
	\subfigure[$t=3.40s$]{\label{fig:d3}\includegraphics[trim=320 30 350 80, clip,width=.24\textwidth]{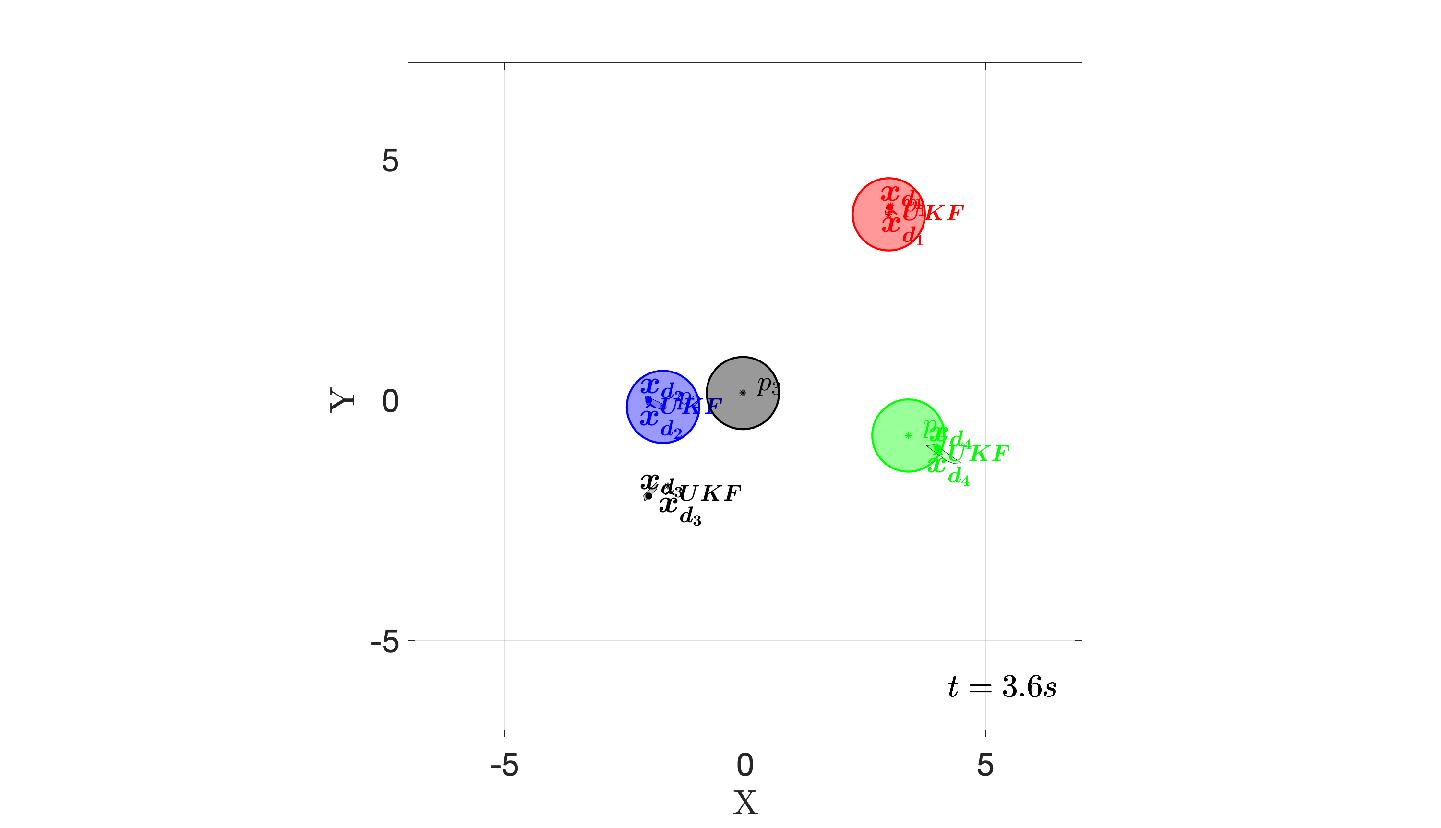}}
	\caption{(a)-(d) Region-based goal identification for multirobot system. Video at \url{https://youtu.be/eh3nApDVUws}}
	\label{MultirobotSnapshots}
\end{figure}

\begin{figure}
	\centering
	\includegraphics[width=\columnwidth]{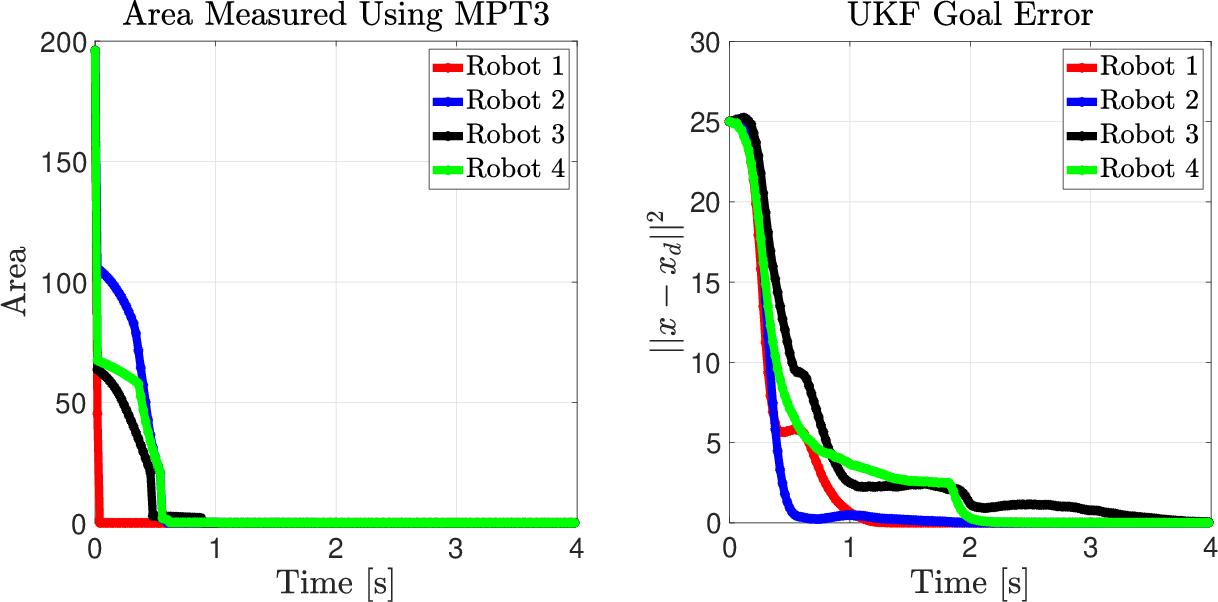}
	\caption{Estimation errors for our proposed region-based estimator(left) and UKF (right). The areas shrink around the goal much faster than UKF errors converge to zero.}
	\label{fig:MultirobotGraphComparison}
\end{figure}

Finally, we consider a multirobot system in Fig. \ref{MultirobotSnapshots} in which we run parallel region-based estimators synchronously. There are four robots shown in different colors, their goals are shown in the same colors as are their feasible regions $\boldsymbol{\Theta}_i(t)$ $\forall i \in \{1,2,3,4\}$. Fig. \ref{fig:MultirobotGraphComparison} compares the convergence of region based estimator (left) to the UKF estimator (right). As is evident, the region-based estimators converge to the true goal much faster than the UKF based estimators.

\section{Conclusions}
In this paper, we proposed a region-based parameter identification method to compute bounds on parameters of tasks being performed by a multirobot system. This approach works even when exact estimation using point-wise estimators is likely to fail. We used KKT conditions to describe the instantaneous feasible set of the underlying parameters using which we computed contracting sets where the parameters must belong. To demonstrate the effectiveness of our method, we showed numerical simulations for inferring goals of a single robot moving amongst static obstacles as well as for each robot in a multirobot system. These scenarios show how our region based estimator outperforms a UKF in terms of speed of convergence. There are several directions that we would like to pursue further. Firstly, we want to see how  physical intervention by the observer be designed to further expedite the convergence of the region-based estimator. Additionally, we also want to explore connections between our work and prior work on inference using inverse convex optimization. 

\label{Conclusions}
\bibliographystyle{ieeetr}
\bibliography{cmu}

\end{document}